\documentclass{article}

\usepackage[preprint,nonatbib]{neurips_2025}

\usepackage[utf8]{inputenc} % allow utf-8 input
\usepackage[T1]{fontenc}    % use 8-bit T1 fonts
\usepackage{url}            % simple URL typesetting
\usepackage{booktabs}       % professional-quality tables
\usepackage{amsfonts}       % blackboard math symbols
\usepackage{microtype}      % microtypography

\usepackage{amsmath}
\usepackage{amssymb}
\usepackage{mathtools}
\usepackage{amsthm}
\usepackage{tensor}
\usepackage{bm}  % assumes amsmath package installed
\usepackage{graphics} % for pdf, bitmapped graphics files
\usepackage{epsfig} % for postscript graphics files
\usepackage{wrapfig}
\usepackage{subcaption}
\usepackage[font=small,labelfont=small]{caption}
\usepackage[pagebackref=true]{hyperref}
\usepackage{color}
\usepackage{algorithm}
\usepackage{algorithmic}
\usepackage{sidecap}
\sidecaptionvpos{figure}{c}
\usepackage{tikz}
\usetikzlibrary{arrows.meta}
\usepackage{tabularx}
\usepackage{colortbl}
\usepackage{booktabs}
\usepackage{multirow}
\usepackage{siunitx}
\usepackage{nicefrac}
\usepackage[nolist]{acronym} % acronyms by the \ac{label} command

\graphicspath{{figures/}}

\definecolor{lightred}{rgb}{0.8, 0.22, 0.29}
\definecolor{turquoise}{rgb}{0.25, 0.89, 0.82}
\definecolor{mediumorchid}{rgb}{0.73, 0.33, 0.83}
\definecolor{darkorange}{rgb}{1.0, 0.65, 0.0}
\definecolor{navy}{rgb}{0.0, 0.0, 0.5}
\definecolor{dodgerblue}{rgb}{0.12, 0.565, 1.0}
\definecolor{crimson}{rgb}{0.86, 0.08, 0.235}
\definecolor{lightgray}{rgb}{0.6, 0.6, 0.6}
\definecolor{darkgreen}{rgb}{0.0, 0.39, 0.0}
\definecolor{middlegreen}{rgb}{0.0, 0.5, 0.0}
\definecolor{middleyellow}{rgb}{1.0, 0.7, 0.0}
\definecolor{aquamarine}{rgb}{0.5, 1.0, 0.83}
\definecolor{lightblue}{rgb}{0.5, 0.7, 1.0}
\definecolor{olive}{rgb}{0.5, 0.5, 0.0}
\definecolor{lightgray}{rgb}{0.6, 0.6, 0.6}

\DeclareRobustCommand{\crimsoncircle}{\tikz{ \filldraw[color=white, fill=crimson, thick](0,0) circle (.075);}}
\DeclareRobustCommand{\darkgreencircle}{\tikz{ \filldraw[color=white, fill=darkgreen, thick](0,0) circle (.075);}}
\DeclareRobustCommand{\middleyellowcircle}{\tikz{ \filldraw[color=white, fill=middleyellow, thick](0,0) circle (.075);}}
\DeclareRobustCommand{\aquamarinecircle}{\tikz{ \filldraw[color=white, fill=aquamarine, thick](0,0) circle (.075);}}
\DeclareRobustCommand{\lightbluecircle}{\tikz{ \filldraw[color=white, fill=lightblue, thick](0,0) circle (.075);}}
\DeclareRobustCommand{\orangecircle}{\tikz{ \filldraw[color=white, fill=darkorange, thick](0,0) circle (.075);}}
\DeclareRobustCommand{\dodgerbluecircle}{\tikz{ \filldraw[color=white, fill=dodgerblue, thick](0,0) circle (.075);}}
\DeclareRobustCommand{\olivecircle}{\tikz{ \filldraw[color=white, fill=olive, thick](0,0) circle (.075);}}
\DeclareRobustCommand{\lightgraycircle}{\tikz{ \filldraw[color=white, fill=lightgray, thick](0,0) circle (.075);}}
\DeclareRobustCommand{\yellowarrow}{\tikz{\draw[color=darkorange, -{Triangle[width = 4pt, length = 2pt]}, line width = 1.2pt] (0.0, 0.0) -- (.5, 0.0);}}
\DeclareRobustCommand{\dodgerbluearrow}{\tikz{\draw[color=dodgerblue, -{Triangle[width = 4pt, length = 2pt]}, line width = 1.2pt] (0.0, 0.0) -- (.5, 0.0);}}
\DeclareRobustCommand{\crimsonline}{\raisebox{2pt}{\tikz{\draw[crimson,solid,line width = 1.1pt](0,0) -- (4mm,0);}}}

\DeclareRobustCommand{\dodgerblueline}{\raisebox{2pt}{\tikz{\draw[dodgerblue,solid,line width = 1.1pt](0,0) -- (4mm,0);}}}

\DeclareRobustCommand{\blackline}{\raisebox{2pt}{\tikz{\draw[black,solid,line width = 1.1pt](0,0) -- (4mm,0);}}}

\DeclareRobustCommand{\blackdashedline}{\raisebox{2pt}{\tikz{\draw[black,dashed,line width = 1.1pt](0,0) -- (4mm,0);}}}
\DeclareRobustCommand{\middlegreendashedline}{\raisebox{2pt}{\tikz{\draw[middlegreen,dashed,line width = 1.1pt](0,0) -- (4mm,0);}}}
\DeclareRobustCommand{\oliveline}{\raisebox{2pt}{\tikz{\draw[olive,solid,line width = 1.1pt](0,0) -- (4mm,0);}}}
\DeclareRobustCommand{\grayline}{\raisebox{2pt}{\tikz{\draw[gray,solid,line width = 1.1pt](0,0) -- (4mm,0);}}}
\DeclareRobustCommand{\verticalblackdashedline}{\raisebox{2pt}{\tikz{\draw[black,dashed,line width = 1.1pt](0,0) -- (0,6mm);}}}
\DeclareRobustCommand{\verticalblackline}{\raisebox{2pt}{\tikz{\draw[black,solid,line width = 1.1pt](0,0) -- (0,6mm);}}}
\DeclareRobustCommand{\verticaldashedsolidblacklinesMNIST}{\raisebox{2pt}{\tikz{\draw[black,dashed,line width = 1.1pt](0,15mm) -- (0,27mm);\draw[black,solid,line width = 1.1pt](0,0) -- (0,12mm);}}}
\DeclareRobustCommand{\verticaldashedsolidblacklinesRobots}{\raisebox{2pt}{\tikz{\draw[black,dashed,line width = 1.1pt](0,13mm) -- (0,25mm);\draw[black,solid,line width = 1.1pt](0,0) -- (0,11mm);}}}
\DeclareRobustCommand{\verticaldashedsolidblacklinesGrasps}{\raisebox{2pt}{\tikz{\draw[black,dashed,line width = 1.1pt](0,14mm) -- (0,24mm);\draw[black,solid,line width = 1.1pt](0,2mm) -- (0,12mm);\draw[white,solid,line width = 1.1pt](0,0mm) -- (0,2mm);}}}

\newcommand{\trsp}{\mathsf{T}}

\DeclareMathOperator*{\argmax}{argmax} % thin space, limits underneath in displays
\DeclareSymbolFont{bbold}{U}{bbold}{m}{n}
\DeclareSymbolFontAlphabet{\mathbbold}{bbold}

\newcommand{\euclideanspace}{\mathbb{R}}
\newcommand{\manifold}{\mathcal{M}}

\newcommand{\norm}[2]{\| #2\|_{#1}}  % Norm of #2 at #1
\newcommand{\expmap}[2]{\text{Exp}_{#1}(#2)}  % Exponential map of #2 at #1
\newcommand{\hyperbolic}[1]{\mathbb{H}^{#1}} % Notation of hyperbolic manifold of dimensionality #1
\newcommand{\grad}{\mathrm{grad}}
\newcommand{\proj}{\mathrm{proj}}

\newcommand{\metric}{\bm{G}}
\newcommand{\lorentzmetric}{\bm{G}^{\mathcal{L}}}
\newcommand{\pullbackmetric}{\bm{G}^{\text{P}}}
\newcommand{\euclpullbackmetric}{\bm{G}^{\text{P},\mathbb{R}}}
\newcommand{\lorentzpullbackmetric}{\bm{G}^{\text{P},\mathcal{L}}}

\newacro{lvm}[LVM]{Latent Variable Model}
\newacro{gplvm}[GPLVM]{Gaussian Process Latent Variable Model}
\newacro{gphlvm}[GPHLVM]{Gaussian Process Hyperbolic Latent Variable Model}

\theoremstyle{plain}

\theoremstyle{definition}

\theoremstyle{remark}

\title{On Probabilistic Pullback Metrics \\for Latent Hyperbolic Manifolds}

\author{%
  Luis Augenstein$^{1}$ \quad\quad No\'emie Jaquier$^{1,2}$ \quad\quad Tamim Asfour$^{1}$ \quad\quad Leonel Rozo$^{3}$\\
  $^{1}$ Karlsruhe Institute of Technology \quad $^{2}$ KTH Royal Institute of Technology \\
  $^{3}$ Bosch Center for Artificial Intelligence \\
}

\begin{document}

\maketitle

\vspace{-0.3cm}
\begin{abstract}
Probabilistic Latent Variable Models (LVMs) excel at modeling complex, high-dimensional data through lower-dimensional representations. Recent advances show that equipping these latent representations with a Riemannian metric unlocks geometry-aware distances and shortest paths that comply with the underlying data structure. This paper focuses on hyperbolic embeddings, a particularly suitable choice for modeling hierarchical relationships. Previous approaches relying on hyperbolic geodesics for interpolating the latent space often generate paths crossing low-data regions, leading to highly uncertain predictions. Instead, we propose augmenting the hyperbolic manifold with a pullback metric to account for distortions introduced by the LVM's nonlinear mapping and provide a complete development for pullback metrics of Gaussian Process LVMs (GPLVMs). Our experiments demonstrate that geodesics on the pullback metric not only respect the geometry of the hyperbolic latent space but also align with the underlying data distribution, significantly reducing uncertainty in predictions.
\end{abstract}

\vspace{-0.35cm}
\section{Introduction}
\vspace{-0.2cm}
\begin{wrapfigure}{r}{0.33\linewidth}
	\vspace{-0.6cm}
	\centering
	\includegraphics[trim={0.2cm 0.2cm 0.5cm 0.2cm},clip,width=0.32\textwidth]{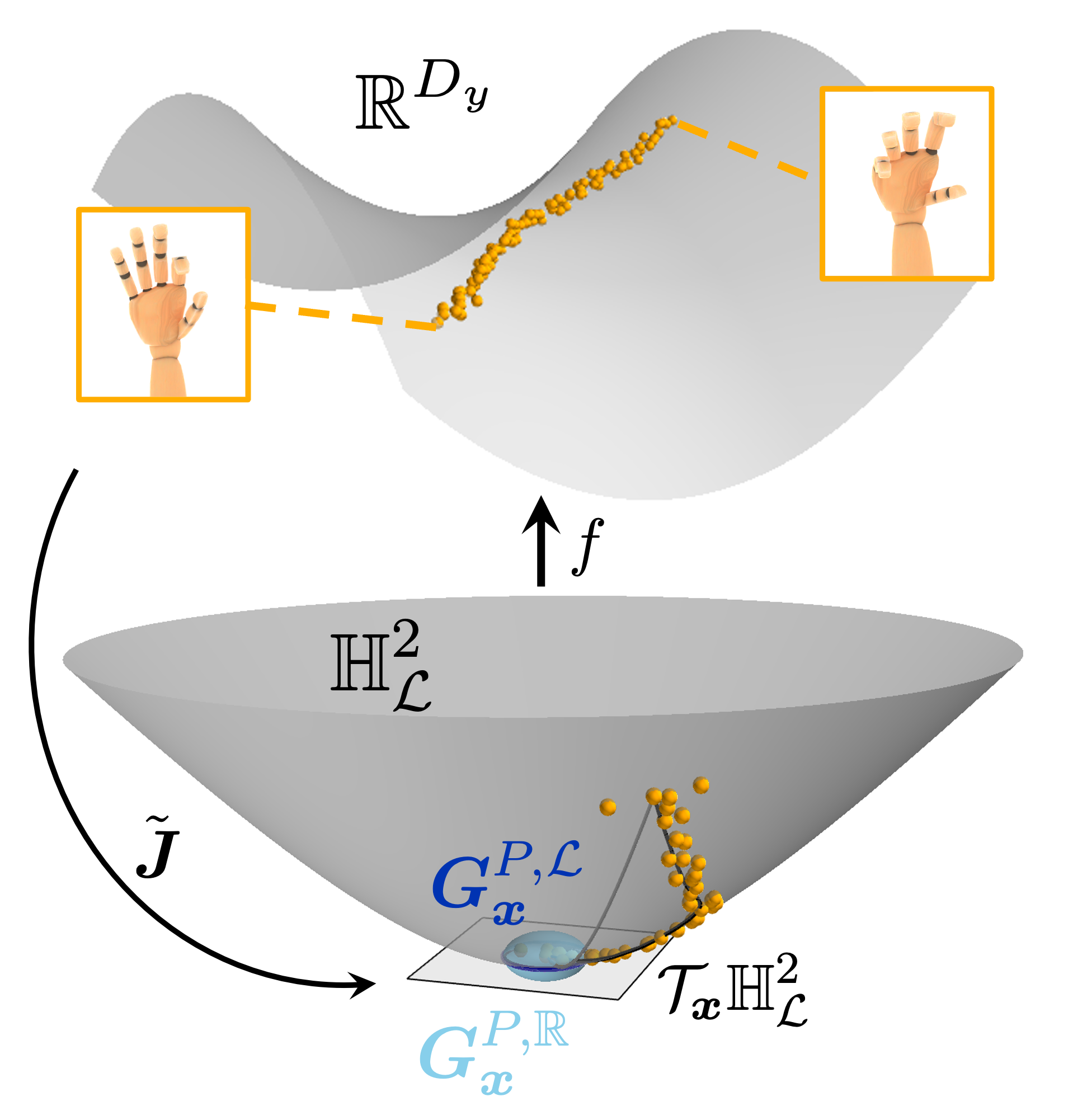}
    \vspace{-0.2cm}
    \caption{Hyperbolic pullback metric on the tangent space of the Lorentz model. The pullback geodesic (\blackline) follows the data (\middleyellowcircle) manifold, in contrast to the hyperbolic geodesic (\grayline).}
	\vspace{-0.4cm}
    \label{fig:concept}
\end{wrapfigure}
Hyperbolic geometry is particularly useful in fields where data exhibits clear hierarchical structures, such as natural language processing for representing word hierarchies and taxonomies~\cite{Nickel2017:Poincare,Nickel2018:Lorentz}, as well as in social network analysis for modeling community structures~\cite{Krioukov10:HyperbolicSocialNetworks,Doorenbos2024:HyperbolicRandomForest}. Additionally, it finds applications in biology~\cite{Alanis-Lobato18:HumanProtein}, human motion taxonomies~\cite{Jaquier2024:GPHLVM}, and computer vision~\cite{Khrulkov20:HyperbolicEmbeddings}. However, many datasets in these disciplines are high dimensional and hetereogenous, making \acp{lvm} indispensable. 
Recently, hyperbolic embeddings gained significant attention due to their ability to capture hierarchical structures inherent in complex high-dimensional data~\cite{Skopek20:MixedCurvVAE,Cho23:HyperbolicVAE,Yang23:HyperbolicRepresentationLearning}. These embeddings leverage hyperbolic growth to accommodate hierarchical relationships, present in trees or graphs, that are difficult to model in Euclidean spaces~\cite{Cvetkovski09:HyperbolicEmbeddigDynGraph}. 
Hyperbolic embeddings excel at preserving the hierarchical relationships of the original data in low-dimensional spaces~\cite{Sala18:HyperbolicRepresentationLearning,Nock2024:HyperbolicEmbeddings}. This can be achieved by learning latent spaces in which embeddings organize according to the data hierarchy~\cite{Mathieu19:HyperbolicVAE,Bose20:HyperbolicNFs}, by leveraging a known taxonomy to guide the embedding process~\cite{Nickel2017:Poincare,Jaquier2024:GPHLVM}, by estimating the original data hierarchy via diffusion geometry~\cite{Lin23:HyperbolicDiffusion}, or by learning a tree on the original data~\cite{Sonthalia20:TreeHyperbolic}. \looseness-1

Despite the potential of hyperbolic embeddings, state-of-the-art techniques operate without accounting for the intrinsic properties of the data. For example, the learned embeddings should be distance-preserving and their distribution should closely match that of the observed data. By doing so, any operation on the hyperbolic latent space complies with the properties of the data manifold. Instead, existing techniques generate geodesics that often traverse low-density regions of the data space, leading to unrealistic interpolations and inaccurate representations.
This problem of integrating the data manifold structure into latent spaces can be tackled via differential geometry~\cite{Hauberg19:OnlyBS}, by leveraging stochastic pullback metrics~\cite{Tosi14:RiemannianGPLVM,Arvanitidis18:LatentSpaceOddity,Arvanitidis21:GeometricallyEnrichedLatentSpaces}. This allows us to introduce uncertainty information into the latent space via a Riemannian metric, enabling the generation of geodesics that faithfully follow the true data distribution.
This has been exploited in robot motion generation~\cite{BeikMohammadi21:GeodesicSkills}, protein sequencing~\cite{Detlefsen22:RiemannianProtein}, and data augmentation for medical imaging~\cite{Chadebec23:DataAugmentationVAE}. 
However, none of the aforementioned works considered hyperbolic geometry as inductive bias on the latent space. \looseness=-1
 
\textbf{This paper} makes several key contributions to the field of hyperbolic \acp{lvm} and \acp{gplvm}. First, we introduce a general formulation of the Riemannian pullback metric on hyperbolic latent spaces, under the assumption of a stochastic latent-to-ambient mapping (see Fig.~\ref{fig:concept}). Second, we present a novel development of the Riemannian pullback metric tailored to \acp{gphlvm}~\cite{Jaquier2024:GPHLVM}, incorporating appropriate Riemannian projections onto tangent spaces to account for the hyperbolic geometry. Third, we develop the kernel derivatives within this setting, highlighting the limitations of current autodifferentiation techniques when applied to our setting. Finally, we demonstrate the benefits of hyperbolic pullback-based geodesics on four experiments: A proof-of-concept $\mathsf{C}$-shape example, MNIST data interpolation, multi-cellular robot design~\cite{Dong24:RobotDesign}, and human grasps generation~\cite{Jaquier2024:GPHLVM}. To the best of our knowledge, this paper is the first to consider pullback metrics of non-Euclidean \acp{lvm}. It sets the ground for further exploitation of Riemannian geometries in latent spaces.

\vspace{-0.2cm}
\section{Background}
\vspace{-0.2cm}
\textbf{Riemannian geometry:}
\label{sec:background-riemannian-geometry}
Before delving into the hyperbolic manifold, it is necessary to establish a basic understanding in Riemannian geometry \cite{Lee18:RiemannManifold}.
A Riemannian manifold $\mathcal{M}$ is a smooth manifold equipped with a Riemannian metric, i.e., a smoothly-varying inner product ${g_{\bm{x}}\!:\mathcal{T}_{\bm{x}}\mathcal{M} \! \times \! \mathcal{T}_{\bm{x}}\mathcal{M} \!\rightarrow \! \mathbb{R}}$ over tangent spaces $\mathcal{T}_{\bm{x}}\mathcal{M}$. When considering coordinates, the Riemannian metric is represented in matrix form for each $\bm{x} \! \in \! \mathcal{M}$ as $\langle \bm{u}, \bm{v} \rangle_{\bm{x}} \!=\! \bm{u}^\trsp \metric_{\bm{x}} \bm{v}$ with $\bm{u},\bm{v}\in\mathcal{T}_{\bm{x}}\mathcal{M}$. The Riemannian metric defines the length of curves in $\mathcal{M}$, leading to the notion of geodesics, defined as locally length-minimizing curves.
To operate with Riemannian manifolds, it is common practice to exploit the Euclidean tangent spaces and the geodesics. 
The exponential map $\text{Exp}_{\bm{x}}(\bm{u}) \!=\! \bm{y}$ maps $\bm{u} \in \mathcal{T}_{\bm{x}}\mathcal{M}$ to a point $\bm{y}\in\mathcal{M}$, so that $\bm{y}$ lies on a geodesic starting at $\bm{x}$ in the direction $\bm{u}$, and such that the geodesic distance $\text{d}_{\mathcal{M}}(\bm{x}, \bm{y})$ equals the length of $\bm{u}$ given by $\Vert \bm{u} \Vert_{\bm{x}} \!=\! \sqrt{\langle \bm{u}, \bm{u} \rangle_{\bm{x}}}$. The inverse operation is the logarithmic map $\text{Log}_{\bm{x}}(\bm{y}) \!=\! \bm{u}$. Finally, the parallel transport $\Gamma_{\bm{x}\rightarrow \bm{y}} (\bm{u}) = \bm{v}$ moves a vector $\bm{u}\in\mathcal{T}_{\bm{x}}\mathcal{M}$ to $\mathcal{T}_{\bm{y}}\mathcal{M}$ while preserving the Riemannian inner product. \looseness-1

Optimizing functions defined on manifold requires generalizing the notion of gradient. The Riemannian gradient $\grad_{\bm{x}}(f)$ of a function $f: \mathcal{M} \to \mathbb{R}$ at $\bm{x}\in\mathcal{M}$ is the unique tangent vector in  $\mathcal{T}_{\bm{x}} \mathcal{M}$ that satisfies $\mathcal{D} _{\bm{u}} f(\bm{x})= \langle \grad_{\bm{x}} f(\bm{x}), \bm{u} \rangle_{\bm{x}}$, where $\mathcal{D}_{\bm{u}} f(\bm{x})$ is the directional derivative of $f$ in the direction $\bm{u}\in\mathcal{T}_{\bm{x}}\mathcal{M}$~\cite[Chap. 3]{Boumal22:RiemannOpt}. The Riemannian Jacobian $\tilde{\bm{J}}$ of a function $f: \mathcal{M} \to \mathbb{R}^{D}$ is composed by the Riemannian gradients for each output dimension, i.e., 
\begin{equation}
    \tilde{\bm{J}} = [\grad_{\bm{x}}(f_1) \:\ldots\: \grad_{\bm{x}}(f_{D})]^\trsp.
    \label{eq:RiemannianJacobian}
\end{equation}

\textbf{Hyperbolic Manifold:}
\label{sec:background-hyperbolic-manifold}
The hyperbolic manifold is the only Riemannian manifold with constant negative curvature~\cite{Ratcliffe19:HyperbolicManifold}. It is commonly represented by either the Poincaré model $\mathbb{H}^{D}_{\mathcal{P}}$~\cite{Poincare00:PoincareModel}, employing local coordinates within the unit ball, or the Lorentz model $\mathbb{H}^{D}_{\mathcal{L}}$~\cite{Jansen09:Hyperboloid,Reynolds93:Hyperboloid}, using Cartesian coordinates to represent the surface embedded in $\mathbb{R}^{D+1}$. In this paper we mostly rely on the latter, which is numerically more stable and defined as 
    $\mathbb{H}^{D}_{\mathcal{L}} = \{ \bm{x} \in \mathbb{R}^{D+1} \mid \langle \bm{x}, \bm{x} \rangle_\mathcal{L} = -1, x_0 > 0 \} $,
where $\langle \bm{x}, \bm{y} \rangle_\mathcal{L} = \bm{x}^\trsp \lorentzmetric \bm{y}$ is the Lorentzian inner product with metric $\lorentzmetric = \text{diag}(-1, 1, ..., 1)$.
Further details on the hyperbolic manifold and its operations are provided in App.~\ref{app-sec:hyperbolic-manifold}.

\textbf{Gaussian Process Hyperbolic Latent Variable Model (GPHLVM):}
\label{sec:background-gphlvm}
A \ac{gplvm} defines a generative mapping from latent variables $\bm{x}_n \!\in\! \mathbb{R}^{D_x}$ to observations $\bm{y}_n\!\in\! \mathbb{R}^{D_y}$ with $D_x < D_y$ through a non-linear transformation modeled by Gaussian processes (GPs)~\cite{Lawrence03:GPLVM}. A \ac{gplvm} assumes that the observations are normally distributed, i.e., $y_{n,d} \sim \mathcal{N}(f_{n,d}, \sigma_y^2)$, and described via a GP and a prior,
\begin{equation}
    f_{n,d} \sim \text{GP}(0, k(\bm{x}_n,\bm{x}_n))  \quad\text{and}\quad \bm{x}_n \sim \mathcal{N}(\bm{0},\bm{I}),
    \label{eq:GPLVM}
\end{equation}
where $y_{n,d}$ is the $d$-th dimension of $\bm{y}_n$, $k:\mathbb{R}^{D_x}\times \mathbb{R}^{D_x} \to \mathbb{R}$ is the GP kernel function which measures the similarity between two inputs $\bm{x}_n,\bm{x}_m$, and $\sigma_y^2$ is the noise variance. 

A Euclidean latent space may not necessarily comply with hierarchical properties of the data, and therefore curved geometries such as the hyperbolic manifold may be preferred.
In such cases, the latent variables $\bm{x}_n \in \mathbb{H}_{\mathcal{L}}^{D_x}$ live in a hyperbolic space and the generative mapping is defined via a \ac{gphlvm}~\cite{Jaquier2024:GPHLVM}. Consequently the kernel function in~\eqref{eq:GPLVM} is replaced by a hyperbolic kernel $k^{\mathbb{H}_{\mathcal{L}}^{D_x}}:\mathbb{H}_{\mathcal{L}}^{D_x}\times \mathbb{H}_{\mathcal{L}}^{D_x} \to \mathbb{R}$ as introduced shortly. 
Moreover, the latent variables are assigned a hyperbolic wrapped Gaussian prior
$\bm{x}_n \sim \mathcal{N}_{\mathbb{H}^{D_x}_{\mathcal{L}}}(\bm{\mu}_0, \alpha \bm{I})$, with $\bm{\mu}_0=(1, 0, \ldots, 0)^\trsp$ and $\alpha$ controlling the spread of the latent variables (see App.~\ref{app-sec:hyperbolic-manifold} for more details). 
The \ac{gphlvm} latent variables and hyperparameters are inferred via MAP or variational inference similar as in the Euclidean case.\looseness=-1 

\textbf{Hyperbolic Kernels:}
The SE and Matérn kernels are standard choices when designing GPs~\cite{Rasmussen06:GPML}. These kernels have been recently generalized to non-Euclidean spaces such as manifolds~\cite{Borovitskiy20:GPManifolds, Jaquier21:GaBOMatern}, or graphs~\cite{Borovitskiy21:GPGraph}. Grigorian and Noguchi~\cite{GrigoryanNoguchi98:HyperbolicHeatKernel} demonstrated that, in hyperbolic space, $2$- and $3$-dimensional SE kernels suffice since higher-dimensional kernels can be expressed as derivatives of these lower-dimensional ones. The 2D and 3D hyperbolic SE kernels are given as,
\begin{equation}
     k^{\mathbb{H}^2_{\mathcal{L}}}(\bm{x}, \bm{z}) = \dfrac{\tau}{C_{\infty}} \int_{\rho}^{\infty} \dfrac{s \, e^{-\frac{s^2}{2\kappa^2}}}{\sqrt{\text{cosh}(s) - \text{cosh}(\rho)}} \text{d} s \, 
     \quad \text{and} \quad
    k^{\mathbb{H}^3_{\mathcal{L}}}(\bm{x}, \bm{z}) = \dfrac{\tau}{C_{\infty}} \dfrac{\rho}{\text{sinh}(\rho)} e^{-\frac{\rho^2}{2\kappa^2}} \, ,
    \label{eq:heat-kernel-2-3D}
\end{equation}
where $\rho = \text{d}_{\mathbb{H}^{D_x}_{\mathcal{L}}}(\bm{x}, \bm{z})$ is the geodesic distance between $\bm{x}, \bm{z} \in \mathbb{H}^{D_x}_{\mathcal{L}}$, $\tau, \kappa \in \mathbb{R}_+$ are the kernel variance and lengthscale, and $C_{\infty}$ is a normalization constant. 
As no closed form solution is known in the $2$D case, the kernel needs to be approximated via a discretization of the integral in~\eqref{eq:heat-kernel-2-3D}. We use the positive semidefinite Monte Carlo approximation introduced in \cite{Jaquier2024:GPHLVM},
\begin{equation}
\label{eq:HypeKernelMonteCarlo}
k^{\mathbb{H}^2_{\mathcal{L}}}(\bm{x}, \bm{z}) \!\approx \dfrac{\tau}{C_{\infty}} \dfrac{1}{L} \sum_{l=1}^{L} s_l\text{tanh}(\pi s_l) \Phi_l(\bm{x}_{\mathcal{P}}, \bm{z}_{\mathcal{P}}) \, ,
\end{equation}
where $\Phi_l(\bm{x}_{\mathcal{P}}, \bm{z}_{\mathcal{P}}) \!=\!  \phi_l(\bm{x}_{\mathcal{P}})\bar{\phi}_l(\bm{z}_{\mathcal{P}})$ is the only component depending on the kernel inputs, $\bar{\phi}_l$ is the complex conjugate of $\phi_l( \bm{x}_{\mathcal{P}}) \!=\!  e^{(1+2s_li)\langle \bm{x}_{\mathcal{P}}, \bm{b}_l \rangle}$, $\bm{x}_{\mathcal{P}}\in \mathbb{H}^{D_x}_{\mathcal{P}}$ is the Poincaré representation of $\bm{x}$, $\bm{b}_l \sim U(\mathbb{S}^1)$ are samples from the unit circle, ${s_l \sim e^{-\frac{s^2\kappa^2}{2}}1_{[0, \infty]}(s)}$ are samples from a truncated Gaussian distribution,  and $\langle \bm{x}_\mathcal{P}, \bm{b} \rangle \!=\! \frac{1}{2} \text{log}\left( \frac{1 - \vert \bm{x}_{\mathcal{P}}\vert^2}{\vert \bm{x}_{\mathcal{P}} - \bm{b} \vert^2} \right)$ is the hyperbolic outer product.

\textbf{Pullback metrics:}
Consider an immersion ${f \colon \mathcal{S}\to \mathcal{M}}$ from a latent space $\mathcal{S}$ to a Riemannian manifold $\mathcal{M}$ equipped with a Riemannian metric $g_{\bm{y}}$. The immersion $f$ induces a pullback metric $g_{\bm{x}}^{\text{P}}$ on $\mathcal{S}$ which, for $\bm{x}\in \mathcal{S}$ and $\bm{u}, \bm{v} \in \mathcal{T}_{\bm{x}}\mathcal{S}$, which is given by~\cite[Chap. 2]{Lee18:RiemannManifold}, 
\begin{equation}
    g^{\text{P}}_{\bm{x}}(\bm{u}, \bm{v}) = g_{f(\bm{x})}\big(d f_{\bm{x}}(\bm{u}), d f_{\bm{x}}(\bm{v})\big). 
    \label{eq:pullback}
\end{equation}
With coordinates, the pullback metric is given in matrix form by $\pullbackmetric_{\bm{x}} = \tilde{\bm{J}}^\trsp \metric_{\bm{y}} \tilde{\bm{J}}$, 
where $\tilde{\bm{J}}$ is the Riemannian Jacobian~\eqref{eq:RiemannianJacobian} of $f$ at $\bm{x}$. Intuitively, $g^{\text{P}}_{\bm{x}}$ evaluates on tangent vectors of $\mathcal{T}_{\bm{x}}\mathcal{S}$ by moving them to $\mathcal{T}_{f(\bm{x})}\mathcal{M}$ to compute their inner product. For an immersion $f:\mathcal{S}\to\mathbb{R}^{D_y}$ with Euclidean co-domain, i.e., $\metric_{\bm{y}} = \bm{I}$, the pullback metric is $\pullbackmetric_{\bm{x}} = \tilde{\bm{J}}^\trsp \tilde{\bm{J}}$. For Euclidean domains $\mathcal{S}=\mathbb{R}^{D_x}$, $\tilde{\bm{J}}$ equals the Euclidean Jacobian $\bm{J}=[\frac{\partial f_1}{\partial \bm{x}} \:\ldots\: \frac{\partial f_{D_y}}{\partial \bm{x}}]^\trsp\in \mathbb{R}^{D_y \times D_x}$. In this paper, we define pullback metrics of hyperbolic \acp{lvm} by explicitly leveraging immersions $f \colon \mathbb{H}^{D_x}\to \mathbb{R}^{D_y}$, where $f$ is a \ac{gphlvm}. \looseness-1

\vspace{-0.2cm}
\section{Metrics of Hyperbolic LVMs}
\vspace{-0.2cm}
Tosi et al.~\cite{Tosi14:RiemannianGPLVM} introduced the pullback metric for Euclidean \acp{gplvm} and its use to compute geodesics that adhere to the data distribution. Here, we extend this approach to hyperbolic LVMs, with \acp{gphlvm} as a special case.
Section~\ref{sec:lvm-hyperbolic-pullback} provides a general formulation of hyperbolic pullback metrics that applies to any hyperbolic \ac{lvm} defined via an immersion $f:\mathbb{H}^{D_x}_{\mathcal{L}}\to\mathbb{R}^{D_y}$, provided that the Riemannian Jacobian of $f$ is available and follows a Gaussian distribution. Section~\ref{sec:gplvm-hyperbolic-pullback} develops the hyperbolic pullback metric for a specific type of hyperbolic \ac{lvm}, i.e., the \ac{gphlvm}, by deriving the probability distribution of the Riemannian Jacobian of the \ac{gphlvm}'s immersion function, which requires kernels and their derivatives. Finally, we compute hyperbolic pullback geodesics in Sec.~\ref{sec:pullback-geodesics}.

\vspace{-0.1cm}
\subsection{A General Hyperbolic Pullback Metric}
\label{sec:lvm-hyperbolic-pullback}
\vspace{-0.1cm}
A deterministic immersion ${f:\mathbb{R}^{D_x}\to\mathbb{R}^{D_y}}$ pulls back the metric into the latent space following~\eqref{eq:pullback}, i.e., ${\euclpullbackmetric_{\bm{x}} = \bm{J}^\trsp \bm{J}}$. 
In the context of \acp{lvm}, the immersion $f$ is stochastic. 
As developed in~\cite{Tosi14:RiemannianGPLVM}, the stochastic immersion $f$ induces a distribution over its Jacobian $\bm{J}$, which itself induces a distribution over the pullback metric. 
As~\cite{Tosi14:RiemannianGPLVM}, we consider \acp{lvm} for which the probability over $\bm{J}$ follows a Gaussian distribution. Assuming independent rows $\bm{J}_d \in \mathbb{R}^{D_x}$, each with its own mean but shared covariance matrix, the Jacobian distribution is of the form,
\begin{equation}
    p(\bm{J}) = \prod_{d=1}^{D_y} \mathcal{N}(\bm{J}_d \mid \bm{\mu}_{\bm{J}_d}, \bm{\Sigma}_J) \, .
    \label{eq:EuclideanJacobianDistribution}
\end{equation}
Therefore, the metric tensor $\euclpullbackmetric_{\bm{x}}$ follows a non-central Wishart distribution~\cite{Anderson46:Wishart},
\begin{equation}
p(\euclpullbackmetric_{\bm{x}}) = \mathcal{\bm{W}}_{D_x}(D_y, \bm{\Sigma}_J, \mathbb{E}[\bm{J}]^\trsp \mathbb{E}[\bm{J}]) \, ,
\label{eq:EuclideanPullbackMetricDistribution}
\end{equation} 
which we compute the expected metric tensor from,
\begin{align}
    \label{eq:euclidean-pullback-metric-tensor}
    \mathbb{E}[\euclpullbackmetric_{\bm{x}}] = \mathbb{E}[\bm{J}]^\trsp \mathbb{E}[\bm{J}] + D_y \bm{\Sigma}_J \, .
\end{align}

While the hyperbolic case follows a similar strategy, special care is required to ensure that the Jacobian rows lie on appropriate tangent spaces. Hyperbolic \acp{lvm} define a stochastic mapping $f: \mathbb{H}^{D_x}_{\mathcal{L}} \rightarrow \mathbb{R}^{D_y}$, whose Jacobian is formed by the Riemannian gradients of $f$ as in~\eqref{eq:RiemannianJacobian}.
As for Riemannian submanifolds~\cite{Boumal22:RiemannOpt}, each Riemannian gradient $\grad_{\bm{x}}(f_d)$ equals the orthogonal projection of the Euclidean gradient $\frac{\partial f_d}{\partial \bm{x}}\in\mathbb{R}^{D_x+1}$ onto $\mathcal{T}_{\bm{x}}\mathbb{H}^{D_x}_{\mathcal{L}}$, 
\begin{equation}
    \grad_{\bm{x}}(f_d) = \proj_{\bm{x}} \left(\frac{\partial f_d}{\partial \bm{x}} \right) = \bm{P}_{\bm{x}} \frac{\partial f_d}{\partial \bm{x}} ,
\end{equation}
with $\bm{P}_{\bm{x}} \!=\! \lorentzmetric + \bm{x}\bm{x}^\trsp$. 
Therefore, the Riemannian Jacobian $\tilde{\bm{J}}$ is computed from the Euclidean Jacobian $\bm{J}$ as ${\tilde{\bm{J}} \!=\! [\bm{P}_{\bm{x}} \frac{\partial f_1}{\partial \bm{x}} \:\ldots\: \bm{P}_{\bm{x}} \frac{\partial f_{D_y}}{\partial \bm{x}}]^\trsp \!=\! \bm{J}\bm{P}^\trsp_{\bm{x}}}$. Assuming independent rows $\tilde{\bm{J}}_d$ as in the Euclidean case, the Jacobian distribution is computed from~\eqref{eq:EuclideanJacobianDistribution} as~\cite[Chap. 2.3]{Bishop06:GaussianProperties},
\begin{equation}
    p(\tilde{\bm{J}}) = \prod_{d=1}^{D_y} \mathcal{N}\big(\tilde{\bm{J}}_d \mid \bm{P}_{\bm{x}} \bm{\mu}_{\bm{J}_d}, \bm{P}_{\bm{x}}\bm{\Sigma}_J\bm{P}_{\bm{x}}^\trsp\big) \, .
    \label{eq:RiemannianJacobianDistribution}
\end{equation}
Similarly to $\euclpullbackmetric_{\bm{x}}$ in~\eqref{eq:EuclideanPullbackMetricDistribution}, the metric tensor $\lorentzpullbackmetric_{\bm{x}}$ follows a non-central Wishart distribution~\cite{Anderson46:Wishart},
\begin{equation}
p(\lorentzpullbackmetric_{\bm{x}}) = \mathcal{\bm{W}}_{D_x}(D_y, \tilde{\bm{\Sigma}}_J, \mathbb{E}[\tilde{\bm{J}}]^\trsp \mathbb{E}[\tilde{\bm{J}}]) \, ,
\label{eq:HypePullbackMetricDistribution}
\end{equation} 
with $\tilde{\bm{\Sigma}}_J\!=\!\bm{P}_{\bm{x}}\bm{\Sigma}_J\bm{P}_{\bm{x}}^\trsp$, leading to the expected metric, 
\begin{equation}
    \mathbb{E}[\lorentzpullbackmetric_{\bm{x}}] = \mathbb{E}[\tilde{\bm{J}}]^\trsp \mathbb{E} [\tilde{\bm{J}}] + D_y \tilde{\bm{\Sigma}}_J 
    = \bm{P}_{\bm{x}} (\mathbb{E}[\bm{J}]^\trsp \mathbb{E}[\bm{J}] + D_y \bm{\Sigma}_J) \bm{P}_{\bm{x}}^\trsp \, .
    \label{eq:hyperbolic-pullback-metric-tensor}
\end{equation}
Intuitively, the pullback metric $\lorentzpullbackmetric_{\bm{x}}$ is the orthogonal projection of the Euclidean pullback metric $\euclpullbackmetric_{\bm{x}}$ onto the tangent space $\mathcal{T}_{\bm{x}}\mathbb{H}^{D_x}_{\mathcal{L}}$ (see Fig.~\ref{fig:concept}). It relates to the hyperbolic metric $\lorentzmetric$ via the projection $\bm{P}_{\bm{x}}$, which is orthogonal with respect to the Lorentzian inner product. Note that the presented approach allows us to obtain pullback metrics for all hyperbolic \acp{lvm} where the Jacobian $\bm{J}$ of $f$ follows a Gaussian distribution, including \acp{gphlvm}~\cite{Jaquier2024:GPHLVM} and hyperbolic VAEs~\cite{Nagano19:HyperbolicNormal,Mathieu19:HyperbolicVAE}.

\vspace{-0.1cm}
\subsection{The \ac{gphlvm} Pullback Metric}
\label{sec:gplvm-hyperbolic-pullback}
\vspace{-0.1cm}
Here, we consider the case where the mapping $f$ is defined as a \ac{gphlvm}. We derive the distribution of the Riemannian Jacobian, which we then use to compute the expected \ac{gphlvm} pullback metric $\lorentzpullbackmetric_{\bm{x}^*}$. 
As for \acp{gplvm}~\cite{Tosi14:RiemannianGPLVM}, the joint distribution of the Euclidean Jacobian $\bm{J}$ and observations $\bm{Y}=[\bm{y}_1 \ldots \bm{y}_N]^\trsp\in\mathbb{R}^{N \times D_y}$ is given as,
\begin{align}
    \label{eq:gplvm-jacobian-joint-distribution}
    \mathcal{N}\left(\begin{bmatrix}
    \bm{Y}_d \\
    \bm{J}_d
    \end{bmatrix} \Bigg| \begin{bmatrix} \bm{0} \\
    \bm{0} \end{bmatrix}, \begin{bmatrix}
    \bm{K}_X + \sigma_y^2\bm{I}_N & \partial k(\bm{X}, \bm{x}^*) \\
     \partial k(\bm{x}^*, \bm{X}) & \partial^2 k(\bm{x}^*, \bm{x}^*)
    \end{bmatrix} \right) ,
\end{align}
where $\bm{X}=[\bm{x}_1 \ldots \bm{x}_N]^\trsp\in\mathbb{R}^{N \times (D_x+1)}$ with $\bm{x}_n \in \mathbb{H}^{D_x}_{\mathcal{L}}$, $\bm{K}_X\in\mathbb{R}^{N \times N}$ is the corresponding kernel matrix, $\bm{Y}_d$ denotes the $d$-th row of $\bm{Y}$, and we rely on the hyperbolic kernel derivatives,
\begin{equation}
    \label{eq:kernel-derivative}
    \partial k(\bm{X}, \bm{x}^*) = \frac{\partial}{\partial \bm{z}} k(\bm{X}, \bm{z}) \Big\vert_{\bm{z}=\bm{x}^*} 
    \quad\quad \text{and} \quad\quad
    \partial^2 k(\bm{x}^*, \bm{x}^*) = \frac{\partial^2}{\partial \bm{z} \, \partial \bm{x}} k(\bm{x}, \bm{z}) \Big\vert_{\bm{x}=\bm{z}=\bm{x}^*} \, .
\end{equation}
While in the Euclidean case the kernel derivatives are straightforward, the hyperbolic setting is more challenging, as will be discussed in Sec.~\ref{sec:hyperbolic-kernel-derivatives}.
Conditioning on the observations $\bm{Y}_d \in \mathbb{R}^{N}$ results in the probability distribution of the Jacobian at $\bm{x}^*$ \cite[Chap 2.3]{Bishop06:GaussianProperties},
\begin{equation}
    \label{eq:GPHLVM-jacobian}
    p(\bm{J} \mid \bm{Y}, \bm{X}, \bm{x}_*) = \prod_{d=1}^{D_y} \mathcal{N}(\bm{J}_d \mid \bm{\mu}_{\bm{J}_d}, \bm{\Sigma}_J) \,\, \text{with} \,\,
    \bm{\mu}_{\bm{J}_d} = \bm{S}_J \bm{Y}_d 
    \,\text{,} \,
    \bm{\Sigma}_J = \partial^2 k(\bm{x}^*, \bm{x}^*) -  \bm{S}_J \partial k(\bm{X}, \bm{x}^*),
\end{equation}
and $\bm{S}_J = \partial k(\bm{x}^*, \bm{X})(\bm{K}_{X} + \sigma_y^2 \bm{I}_N)^{-1}$. The distribution of the Riemannian Jacobian $\tilde{\bm{J}}$ is then obtained from~\eqref{eq:GPHLVM-jacobian} by using~\eqref{eq:RiemannianJacobianDistribution}.
The expected pullback metric is finally computed using~\eqref{eq:hyperbolic-pullback-metric-tensor} as,
\begin{align}
    \label{eq:gphlvm-expected-pullback-metric}
    \mathbb{E}[\lorentzpullbackmetric_{\bm{x}^*}] = \bm{P}_{\bm{x}^*}(\bm{\mu}_J^\trsp \bm{\mu}_J + D_y \bm{\Sigma}_J) \bm{P}_{\bm{x}^*}^\trsp \, , \quad \text{with} \quad \bm{\mu}_J = [\bm{\mu}_{\bm{J}_1} \: \ldots \: \bm{\mu}_{\bm{J}_{D_y}}]^\trsp \in \mathbb{R}^{D_y \times D_x} \, . 
\end{align}

\subsection{Hyperbolic Pullback Geodesics}
\label{sec:pullback-geodesics}
\begin{figure}[tbp]
    \centering
    \includegraphics[width=.95\textwidth]{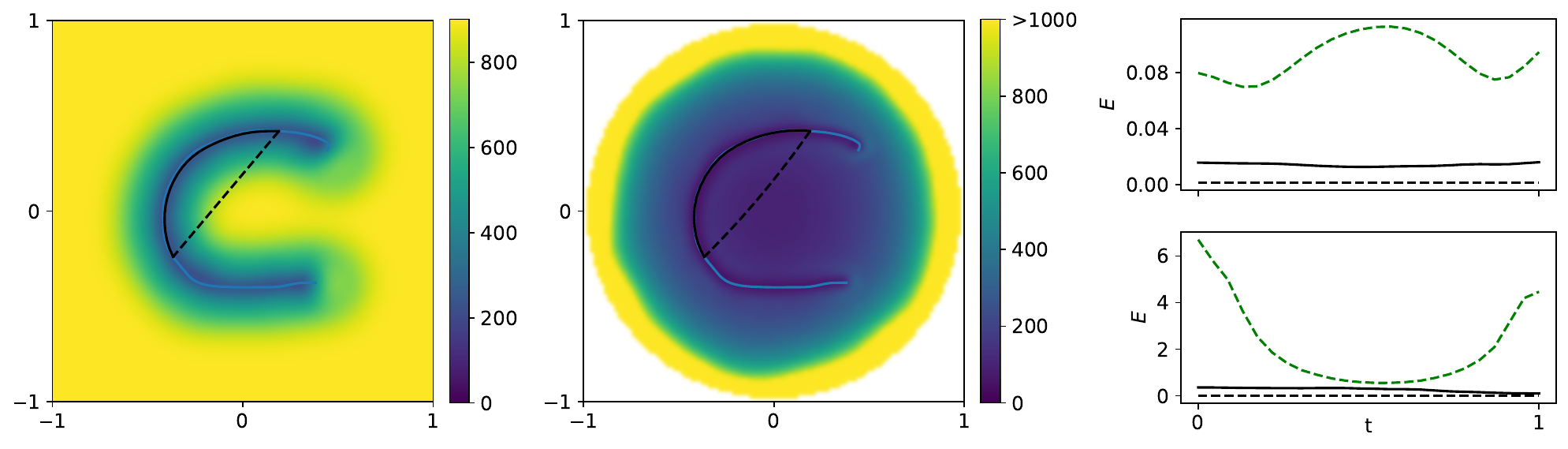}
    \caption{\emph{Left} and \emph{middle}: Euclidean and hyperbolic pullback metrics on an embedded $\mathsf{C}$-shape trajectory (\dodgerblueline) with base manifold (\blackdashedline) and pullback (\blackline) geodesics. \emph{Right}: Curve energy along the geodesics on the Euclidean (top) and hyperbolic (bottom) cases. These include the base manifold geodesic (\blackdashedline), the base manifold geodesic with energy~\eqref{eq:curve_energy} evaluated using the pullback metric (\middlegreendashedline), and the pullback geodesic (\blackline).}
    \vspace{-0.4cm}
    \label{fig:c-shape}
\end{figure}

Vanilla geodesics on hyperbolic latent spaces naturally adhere to the hyperbolic geometry. However, they do not account for the underlying data structure and can traverse sparse data regions. To overcome this, we leverage pullback metrics on the hyperbolic manifold to design geodesics that comply with the geometry of both the hyperbolic space and the data distribution (see Figs.~\ref{fig:concept} and~\ref{fig:c-shape}). 
Unlike hyperbolic geodesics that have a closed-form solution, hyperbolic pullback geodesics require solving an optimization problem. 
Specifically, pullback geodesics are computed by minimizing the curve length, or equivalently the curve energy $E$ with respect to the pullback metric. Considering a discretized geodesic composed by a set of $M$ points $\bm{x}_i\in \mathbb{H}^{D_x}_{\mathcal{L}}$, this boils down to minimize,
\begin{align}
    \label{eq:curve_energy}
    E = \sum_{i=0}^{M-2} \bm{v}_i^\trsp \lorentzpullbackmetric_{\bm{x}_i} \bm{v}_i \, , \quad \text{with} \quad \bm{v}_i = \text{Log}_{\bm{x}_i}({\bm{x}_{i+1}}).
\end{align}
While it is possible to iteratively optimize the curve points $\bm{x}_i$ directly, this often leads to uneven spacing among them. This issue can be addressed either by using a parametric curve on the manifold~\cite{Gousenbourger14:RiemannianBézierCurves} and optimize its parameters instead of the points directly, or by incorporating the spline energy as a regularization factor in the optimization process~\cite{Heeren17:RiemannianSplines}. We follow the latter. The spline energy is  
$E_{\text{spline}} \!\approx\! \sum_{i=1}^{M-2} \text{d}_{\mathbb{H}^{D_x}_{\mathcal{L}}}(\bm{x}_i, \bar{\bm{x}}_i)^2$, 
where $\bar{\bm{x}}_i = \text{Exp}_{\bm{x}_{i-1}}\left(\frac{1}{2} \, \text{Log}_{\bm{x}_{i-1}}(\bm{x}_{i+1}) \right)$ is the geodesic midpoint between $\bm{x}_{i-1}$ and $\bm{x}_{i+1}$. The final optimization problem is, 
\begin{equation}
\label{eq:loss_energy_regularized}
\min_{\bm{x}_0, ..., \bm{x}_{M-1}} E + \lambda E_{\text{spline}}, 
\end{equation}
with $\lambda$ weighting the influence of the regularization. As the optimization parameters are Riemannian, we leverage Riemannian optimizers such as Riemannian Adam~\cite{Becigneul19:RiemannianAdaptiveOpt} to optimize~\eqref{eq:curve_energy}. The computation of pullback geodesics is summarized in Algorithm~\ref{alg:hyperbolic-pullback-geodesics} in App.~\ref{app:pullback-geodesics}. \looseness=-1

\vspace{-0.2cm}
\section{Hyperbolic Kernel Derivatives}
\label{sec:hyperbolic-kernel-derivatives}
\vspace{-0.2cm}
Here we discuss the challenges of computing derivatives for hyperbolic SE kernels for which standard autodiff tools cannot be used directly. Table~\ref{tab:kernel-derivative-comparison} shows that autodiff-based kernel derivatives computed with PyTorch~\cite{Paszke19:Pytorch} are approximately five times slower than analytic implementations in the 2D case. In the 3D case, autodiff fails to compute the derivatives entirely. To overcome these issues, we derive  analytic kernel derivatives and focus our analysis on the 2D and 3D cases. This choice is motivated 
\begin{wraptable}{r}{6.2cm}
    \vspace{-0.0cm}
    \caption{Average computation times in seconds over $100$ runs, each evaluating the pullback metric $\lorentzpullbackmetric_{\bm{x}}$ and its derivatives at a random point $\bm{x}$. %
    }
    \vspace{-0.3cm}
    \label{tab:kernel-derivative-comparison}
    \begin{center}
    \resizebox{6.0cm}{!}{
    \begin{tabular}{ccc}
    \toprule
        & $k^{\mathbb{H}^2_{\mathcal{L}}}$ ($L=3000)$~\eqref{eq:HypeKernelMonteCarlo}      & $k^{\mathbb{H}^3_{\mathcal{L}}}$~\eqref{eq:heat-kernel-2-3D}   \\
        \midrule
        PyTorch         & $0.83 \pm 0.06$ & $-$             \\
        Analytic      & $\bm{0.16 \pm 0.02}$ & $\bm{0.06 \pm 0.01}$ \\
    \bottomrule
    \end{tabular}
    }
    \end{center}
    \vspace{-0.6cm}
\end{wraptable}
by the increased capacity of the hyperbolic manifold to embed hierarchical data in low-dimensional spaces, coupled with the computational efficiency and ease of visualization of 2D and 3D latent spaces. Note that the forthcoming derivations could serve as a basis for higher-dimensional hyperbolic kernels defined as derivatives of 2D and 3D kernels~\eqref{eq:heat-kernel-2-3D}~\cite{GrigoryanNoguchi98:HyperbolicHeatKernel}.

\textbf{2D Hyperbolic SE Kernel Derivatives: }
\label{sec:2D-hyperbolic-heat-kernel}
We compute the 2D hyperbolic SE kernel $k^{\mathbb{H}^2_{\mathcal{L}}}(\bm{x}, \bm{z})$ via the Monte Carlo approximation~\eqref{eq:HypeKernelMonteCarlo}. The pullback metric tensor $\lorentzpullbackmetric_{\bm{x}^*}$ is computed in~\eqref{eq:gphlvm-expected-pullback-metric} using the kernel derivatives $\frac{\partial k(\bm{x}, \bm{z})}{\partial \bm{x}} \vert_{\bm{x} = \bm{x}^*}$ and $\frac{\partial^2 k(\bm{x}, \bm{z})}{\partial \bm{z} \partial \bm{x}} \vert_{\bm{x}=\bm{z}=\bm{x}^*}$~\eqref{eq:kernel-derivative}. The only part of the kernel~\eqref{eq:HypeKernelMonteCarlo} that depends on the inputs $\bm{x},\bm{z}$ is the function $\Phi_l(\bm{x}, \bm{z})$, whose derivatives are given in App.~\ref{app-subsec:2Dhyperbolic-kernel-derivatives}. 
Minimizing the curve energy~\eqref{eq:curve_energy} to compute pullback geodesics additionally requires the derivative $\frac{\partial}{\partial \bm{x}^*} \lorentzpullbackmetric_{ \bm{x}^*}$. This, in turn, requires the derivative of the Jacobian of the covariance matrix $\frac{\partial \bm{\Sigma}_J}{\partial \bm{x}^*}$, which depends on the kernel derivatives $\frac{\partial^3 k(\bm{x}, \bm{z})}{\partial \bm{x} \partial \bm{z} \partial \bm{x}} \vert_{\bm{x}=\bm{z}=\bm{x}^*}$, and $\frac{\partial^3 k(\bm{x}, \bm{z})}{\partial \bm{z}^2 \partial \bm{x}}  \vert_{\bm{x}=\bm{z}=\bm{x}^*}$. Again, the kernel derivatives are determined by the derivatives of $\Phi_l(\bm{x}, \bm{z})$. The complete derivation of all derivatives is provided in App.~\ref{app-subsec:2Dhyperbolic-kernel-derivatives}. \looseness-1

\textbf{3D Hyperbolic SE Kernel Derivatives: }
\label{sec:3D-hyperbolic-heat-kernel}
The 3D hyperbolic SE kernel is given by~\eqref{eq:heat-kernel-2-3D}. As in the 2D case, to obtain the expected pullback metric tensor~\eqref{eq:gphlvm-expected-pullback-metric}, we compute the first two kernel derivatives $\frac{\partial k(\bm{x}, \bm{z})}{\partial \bm{x}} \vert_{\bm{x} = \bm{x}^*}$ and $\frac{\partial^2 k(\bm{x}, \bm{z})}{\partial \bm{z} \partial \bm{x}} \vert_{\bm{x}=\bm{z}=\bm{x}^*}$. As derived in App.~\ref{app-subsec:3Dhyperbolic-kernel-derivatives}, these derivatives depend on the function $g(u) \!=\! \left(\frac{2\rho^2}{\nu s^2} - \frac{1}{s^2} - \frac{u\rho}{s^3} \right)$ with $\rho = \text{d}_{\mathbb{H}^{D_x}_{\mathcal{L}}}(\bm{x}, \bm{z})$, $u = \langle \bm{x}, \bm{z} \rangle_{\mathcal{L}}$, and $s = \sqrt{u^2-1}$. The function $g(u)$ is essential to understand why standard automatic differentiation tools fail to compute the 3D kernel derivatives. For equal inputs $\bm{x} \!=\! \bm{z}$, the inner product $u$ approaches $-1$, while the distance $\rho = \text{d}_{\mathbb{H}^{D_x}_{\mathcal{L}}}(\bm{x}, \bm{z})$ and variable $s$ converge to $0$. This limit is analytically well behaved, i.e., $\lim_{u \rightarrow -1^-} g(u) = \frac{2}{\nu} + \frac{1}{3}$. However, automatic differentiation tools fail to compute the kernel derivatives for equal inputs as they do not cancel out $0$-approaching terms. For example, we have,
\begin{equation}
    g(u) = \dfrac{2\rho^2}{\nu s^2} - \dfrac{1}{s^2} - \dfrac{u\rho}{s^3} \rightarrow^{\text{ autodiff}}_{\: \: \: \:\bm{x}=\bm{z}} \: \rightarrow \dfrac{0}{0} - \dfrac{1}{0} - \dfrac{0}{0} = \text{NaN} \, ,
\end{equation}
where divisions by $0$ lead to undefined values (NaN). While derivatives for equal kernel inputs $\bm{x}=\bm{z}$ may be less relevant when training a \ac{gplvm} as different latent points rarely become so close, they are essential for computing the Jacobian covariance matrix $\bm{\Sigma}_J$ in~\eqref{eq:GPHLVM-jacobian}. 
Although symbolic differentiation libraries could provide a solution, they tend to be too slow for practical purposes. Instead, we address this issue by computing the necessary derivatives and their analytical limits manually when the inputs $\bm{x}, \bm{z}$ are closer than a predefined threshold.
Complete derivations are provided in App.~\ref{app-subsec:3Dhyperbolic-kernel-derivatives}. \looseness=-1

\vspace{-0.2cm}
\section{Experiments}
\label{sec:experiments}
\vspace{-0.2cm}
We test the proposed pullback metric in four distinct experiments to demonstrate that \emph{(1)} the hyperbolic pullback metric augments the hyperbolic metric with the distortions introduced by the \ac{gphlvm}'s nonlinear mapping; and \emph{(2)} the hyperbolic pullback geodesics adhere to the data distribution, leading to low uncertainty model predictions. Note that a direct comparison against other state-of-the-art hyperbolic \acp{lvm}~\cite{Mathieu19:HyperbolicVAE,Bose20:HyperbolicNFs} is not feasible at this time, as pullback metrics were not derived yet for these models. Additional experimental details are provided in App.~\ref{app-sec:experimental-details}.\looseness=-1

\textbf{C-shape: }
The first experiment serves as a proof of concept to visualize and compare the Euclidean pullback metric $\euclpullbackmetric$ from a \ac{gplvm} and the hyperbolic pullback metric $\lorentzpullbackmetric$ from a \ac{gphlvm}. We design a dataset of 2D $\mathsf{C}$-shape points as both latent variables and observations. In the hyperbolic case, the latent variables and observations are encoded in the Lorentz model, i.e, $\bm{x},\bm{y} \in \hyperbolic{2}_\mathcal{L}$, but visualized in the Poincaré model for a more intuitive understanding. We fully specify the \acp{lvm} by setting the variance, length scale, and noise variance as $(\tau, \kappa, \sigma_y^2) = (0.7, 0.15, 0.69)$.

Fig.~\ref{fig:c-shape} displays the Euclidean and hyperbolic pullback metric volumes $\sqrt{\text{det}(\euclpullbackmetric)}$ and $\sqrt{\text{det}(\lorentzpullbackmetric)}$. 
In the hyperbolic case, the pullback metric is a $3\times 3$ matrix lying in the 2D tangent space $\mathcal{T}_{\bm{x}^*}\hyperbolic{2}_\mathcal{L}$. Therefore, one of its eigenvalues is always zero, which we exclude to effectively visualize the volume. As shown in Fig.~\ref{fig:c-shape}-\emph{left}, the Euclidean pullback metric volume is small near the C-shape data and increases away from it, until a maximum value is reached. The hyperbolic pullback metric volume is also low nearby the data and additionally follows the hyperbolic geometry: It is low near the origin and increases when moving outwards until it becomes infinite at the boundary of the unit circle (see Fig.~\ref{fig:c-shape}-\emph{middle}). Overall, the hyperbolic pullback metric effectively integrates the properties of both the hyperbolic manifold and the data manifold.

Fig.~\ref{fig:c-shape} also depicts geodesics generated with and without the pullback metric on both latent manifolds and the curve energy~\eqref{eq:curve_energy} along the geodesics. In both cases, the pullback geodesic closely adheres to the training data, thus highlighting the successful adaptation of the pullback approach to hyperbolic spaces. However, a key difference lies in the curve energy cost of traversing the center, which is substantially lower in the hyperbolic space. 
Note that both the base manifold geodesics and the pullback geodesics exhibit constant energy, confirming that the optimized curves are true geodesics on their respective manifolds. For comparison, we also show the curve energy of the base manifold geodesic, but evaluated using the pullback metric tensor. The non-constant energy indicates that the base manifold geodesic is not a true geodesic under the pullback metric.

\begin{figure}[tbp]
    \centering
    \begin{subfigure}[b]{0.495\textwidth}
        \centering
        \includegraphics[trim={0.0cm 0.3cm 18.0cm 0.0cm},clip,width=.545\textwidth]{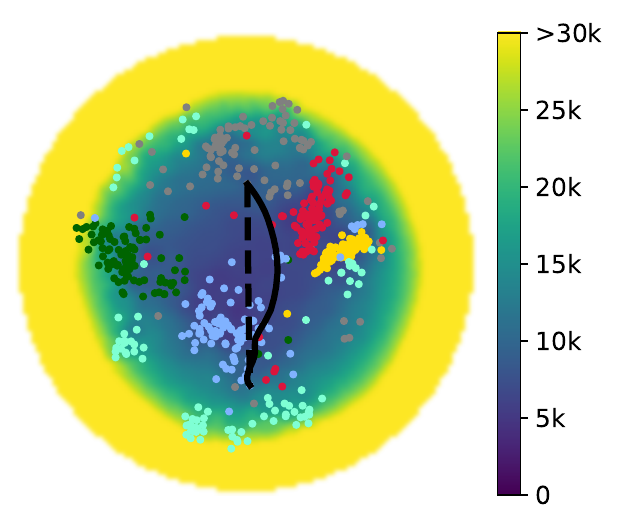}
        \verticaldashedsolidblacklinesMNIST
        \includegraphics[trim={20.0cm 0.35cm 0.0cm 0.0cm},clip,width=.435\textwidth]{figures/mnist_2D.pdf}
        \caption{GPHLVM, $\hyperbolic{2}_\mathcal{L}$}
        \label{fig:mnist_H2}
    \end{subfigure}
    \begin{subfigure}[b]{0.495\textwidth}
        \centering
        \includegraphics[trim={0.0cm 0.3cm 18.0cm 0.0cm},clip,width=.545\textwidth]{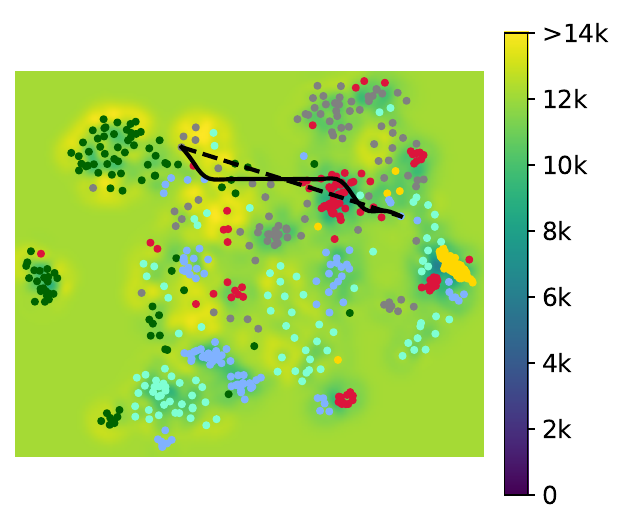}
        \verticaldashedsolidblacklinesMNIST
        \includegraphics[trim={20.0cm 0.35cm 0.0cm 0.0cm},clip,width=.435\textwidth]{figures/mnist_2D_euclidean.pdf}
        \caption{GPLVM, $\euclideanspace^{2}$}
        \label{fig:mnist_R2}
    \end{subfigure}
    \caption{\emph{Left}: Embeddings of a subset of the MNIST dataset with digits $\mathsf{0}$ (\darkgreencircle), $\mathsf{1}$ (\middleyellowcircle), $\mathsf{2}$ (\aquamarinecircle), $\mathsf{3}$ (\lightgraycircle), $\mathsf{6}$ (\lightbluecircle), and $\mathsf{9}$~(\crimsoncircle). The background color represents the pullback metric volume. The base manifold (\blackdashedline) and pullback (\blackline) geodesics interpolate between a $\mathsf{3}$ and a $\mathsf{6}$. 
    \emph{Right}: Ten samples along the decoded geodesics in image space.}
    \vspace{-0.5cm}
    \label{fig:mnist_2D}
\end{figure}

\textbf{MNIST Digits: }
The natural clustering of MNIST images can be viewed as a hierarchy whose nodes are each of the $10$ digit classes. Following this point of view, Mathieu et al.~\cite{Mathieu19:HyperbolicVAE} showed that hyperbolic \acp{lvm} are better suited than their Euclidean counterparts for embedding MNIST images. 
Similarly to previous works~\cite{Arvanitidis18:LatentSpaceOddity,Jorgensen21:IsoGPLVM,Lalchand22:GPLVMstochasticVI}, we here explore the interpolation of handwritten digits from a subset of the MNIST dataset. We embed $600$ vectorized $28\!\times\!28$ images of $6$ classes into 2D hyperbolic and Euclidean latent spaces using \ac{gphlvm} and \ac{gplvm}, and then use the induced pullback metrics.\looseness-1

Fig.~\ref{fig:mnist_2D} shows the learned latent spaces along with a base manifold and a pullback geodesic interpolating between a digit $\mathsf{3}$ and a digit $\mathsf{6}$. The decoded geodesics show the result of the interpolation in the image space. 
As shown in Fig.~\ref{fig:mnist_H2}, the \ac{gphlvm} embeddings tend to be grouped by digits.
As expected, the \ac{gphlvm} predictions collapse to the non-informative GP mean in regions with sparse data, resulting in blurry images. This is particularly evident along the hyperbolic geodesic for $t=[0.11,0.44]$. In contrast, the hyperbolic pullback geodesic tends to avoid sparse data regions by bending towards the red cluster (digit $\mathsf{9}$), leading to less blurry predictions overall.
As shown in Fig.~\ref{fig:mnist_R2}, the \ac{gplvm} embeddings are not as clearly organized as in the hyperbolic latent space and form several groups per digit. As such, despite that the Euclidean pullback geodesic also adheres to the data distribution, the resulting interpolation oscillates between digits.
A quantitative comparison of the \ac{gphlvm} and \ac{gplvm} likelihood values confirms that a hyperbolic geometry is better suited for this dataset (see Table~\ref{tab:experiments}). Moreover, the prediction uncertainty values confirm that pullback metrics better captures the data manifold geometry for both models. This validates the effectiveness of hyperbolic pullback geodesics, which thus may be used, e.g., to measure manifold-aware latent distances and probabilities.\looseness=-1 

\textbf{Multi--cellular Robot Design: }
\begin{figure}[t]
    \centering
    \begin{subfigure}[b]{0.495\textwidth}
        \centering
        \includegraphics[trim={0.0cm 9.7cm 0.0cm 0.0cm},clip,width=0.47\textwidth]{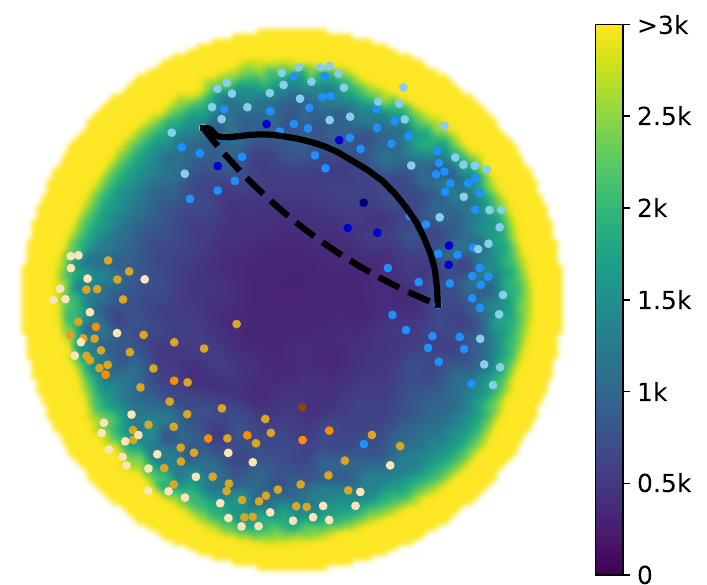}
        \verticaldashedsolidblacklinesRobots
        \includegraphics[trim={0.0cm 0.0cm 0.0cm 10.0cm},clip,width=0.49\textwidth]{figures/multicell_2D.pdf}
        \caption{GPHLVM, $\hyperbolic{2}_\mathcal{L}$}
        \label{fig:multicell_H2}
    \end{subfigure}
    \begin{subfigure}[b]{0.495\textwidth}
        \centering
        \includegraphics[trim={0.0cm 9.7cm 0.0cm 0.0cm},clip,width=0.47\textwidth]{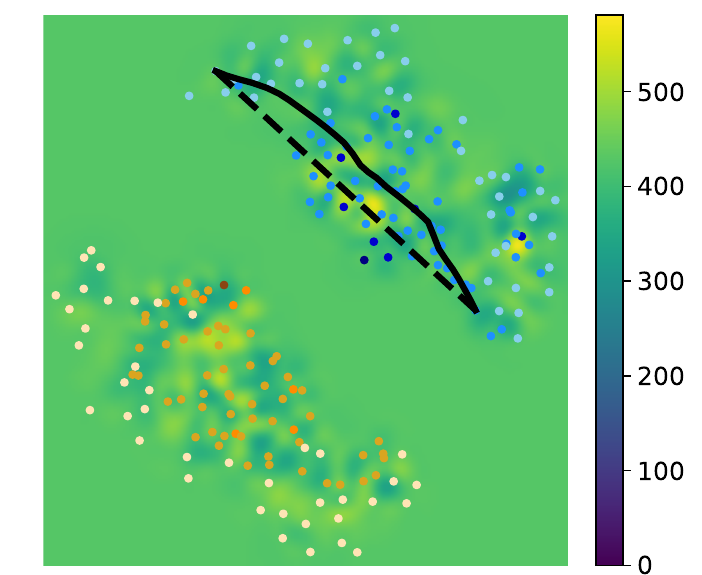}
        \verticaldashedsolidblacklinesRobots
        \includegraphics[trim={0.0cm 0.0cm 0.0cm 10.0cm},clip,width=0.49\textwidth]{figures/multicell_2D_euclidean.pdf}
        \caption{GPLVM, $\euclideanspace^{2}$}
        \label{fig:multicell_R2}
    \end{subfigure}
    \caption{\emph{Left}: Embeddings of multi-cellular robots from coarse (darker tone) to fine (lighter tone). The embeddings form two clusters originating from an all-vertically-actuated robot (\dodgerbluecircle) and an all-horizontally-actuated robot (\orangecircle). \emph{Right}: Ten samples along the decoded base manifold (\blackdashedline) and pullback (\blackline) geodesics.}
    \vspace{-0.5cm}
    \label{fig:multicell_2D}
\end{figure}
We build the coarse--to--fine framework for designing multi--cellular robots using hyperbolic embeddings as introduced by Dong et al.~\cite{Dong24:RobotDesign}. Each robot is composed of a $5 \times 5$ grid of cells, where each cell can be horizontally actuated (orange), vertically actuated (blue), rigid (black), soft (gray), or empty (white). The design process begins with a fully horizontally or vertically actuated robot and incrementally introduces changes, forming a hierarchical structure that refines from coarse to fine architectures. This hierarchy can then be queried to find suitable robots for specific tasks. Dong et al.~\cite{Dong24:RobotDesign} showed that their multi-cellular robot design approach in hyperbolic space outperformed its Euclidean counterpart in most of the considered design tasks.

The main goal of this experiment is to leverage pullback geodesics as a data augmentation mechanism, so that we can design new robot architectures by decoding geodesics that follow the pattern of previously--designed robots.
To do so, we embed existing robot designs into 2D latent spaces using \ac{gphlvm} and \ac{gplvm}. These models are trained with a taxonomy graph-distance prior in the form of a stress loss as in~\cite{Jaquier2024:GPHLVM}, to preserve the hierarchical structure of the designs in the latent space (see App.~\ref{app-sec:multicellular} for details). Then, we compute the pullback metric and optimize geodesics to interpolate between existing robot designs, generating novel ones. Fig.~\ref{fig:multicell_2D} shows the latent space and the mean predictions along both the base manifold and pullback geodesics for each model. As shown in Table~\ref{tab:experiments}, \ac{gphlvm} achieves a lower stress than \ac{gplvm}, indicating that the hyperbolic embeddings better preserve the original hierarchy. 
For both models, the base manifold geodesics cross sparse data regions, while the pullback geodesics adhere to the data support resulting in low-uncertainty predictions (see Fig.~\ref{fig:multicell_2D} and Table~\ref{tab:experiments}). The superior hierarchy preservation in hyperbolic space yields improved transitions between designs, with robot shapes and cell types evolving smoothly along the decoded geodesic (see Fig.~\ref{fig:multicell_H2}). In contrast, the decoded Euclidean geodesic leads to abrupt design changes (see Fig.~\ref{fig:multicell_R2}).
This suggests that hyperbolic pullback geodesics can assist robot design processes.\looseness=-1  

\textbf{Hand Grasps Generation: }
\label{sec:hand-grasps-generation}
\begin{figure*}[t]
    \centering
    \begin{subfigure}[b]{0.495\textwidth}
    \centering
        \includegraphics[trim={1.5cm 8.5cm 14.4cm 0.0cm},clip,width=0.51\linewidth]{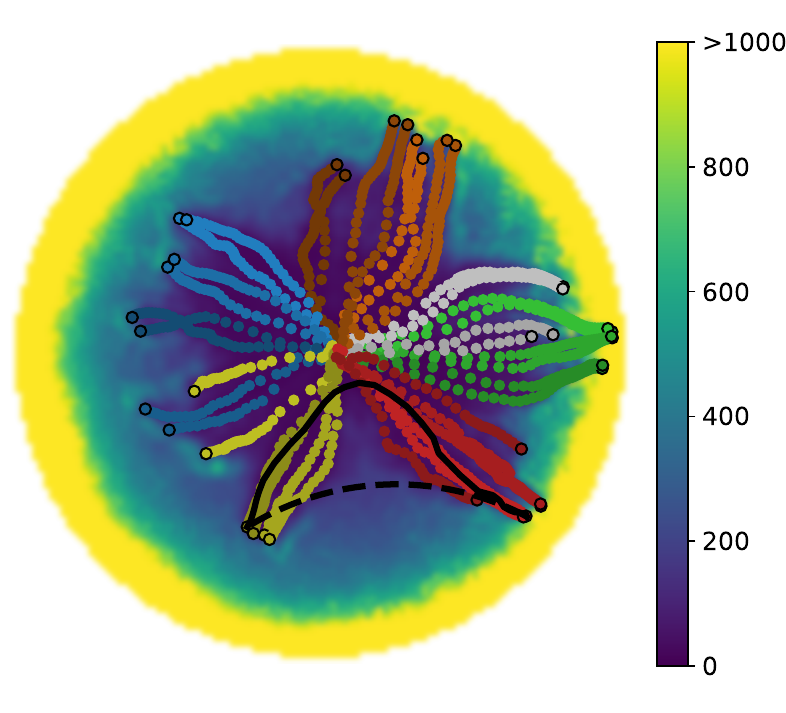}
        \verticaldashedsolidblacklinesGrasps
        \includegraphics[trim={0.1cm 0.0cm 19.5cm 12.0cm},clip,width=0.47\linewidth]{figures/hand_grasps.pdf}
        \verticalblackdashedline
        \includegraphics[trim={0.0cm 0.2cm 0.cm 0.2cm},clip,width=0.95\linewidth]{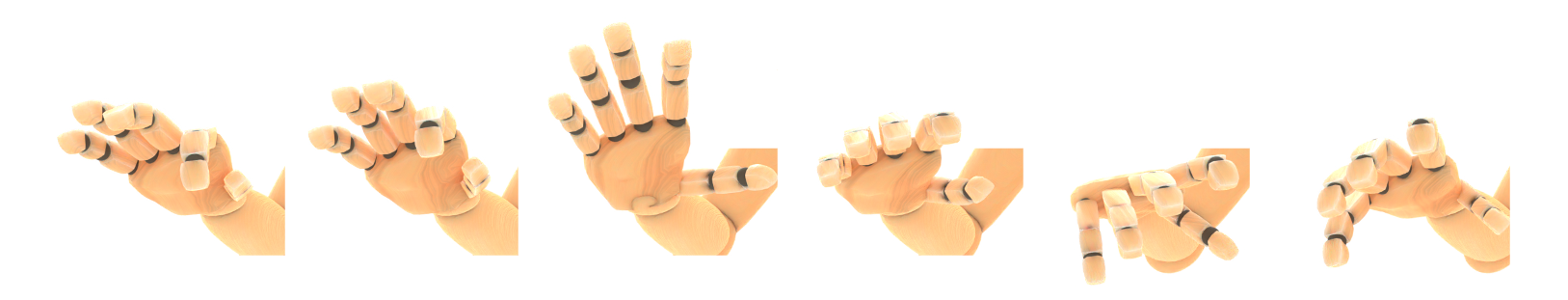}
        \verticalblackline
        \includegraphics[trim={0.0cm 0.2cm 0.cm 0.2cm},clip,width=0.95\linewidth]{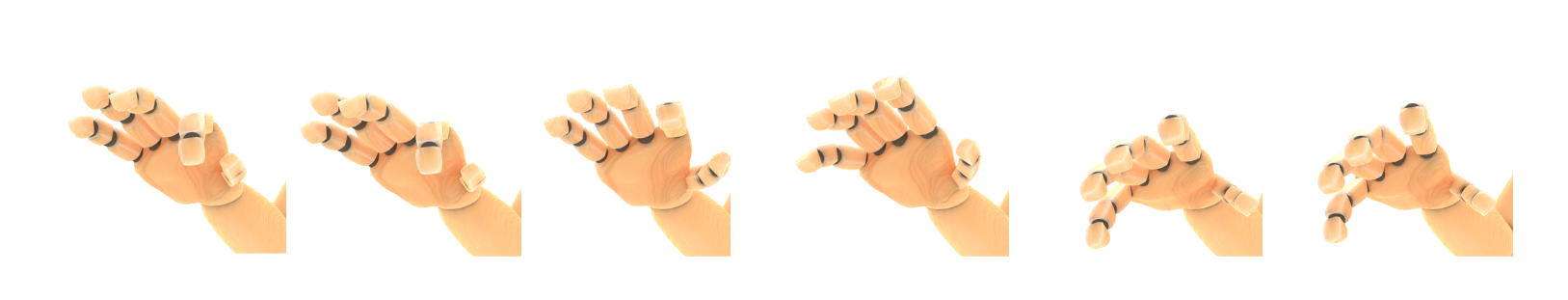}
        \caption{GPHLVM, $\hyperbolic{2}_\mathcal{L}$}
        \label{fig:hand_grasps_H2}
    \end{subfigure}
        \begin{subfigure}[b]{0.495\textwidth}
    \centering
        \includegraphics[trim={1.5cm 9.2cm 14.4cm 0.0cm},clip,width=0.51\linewidth]{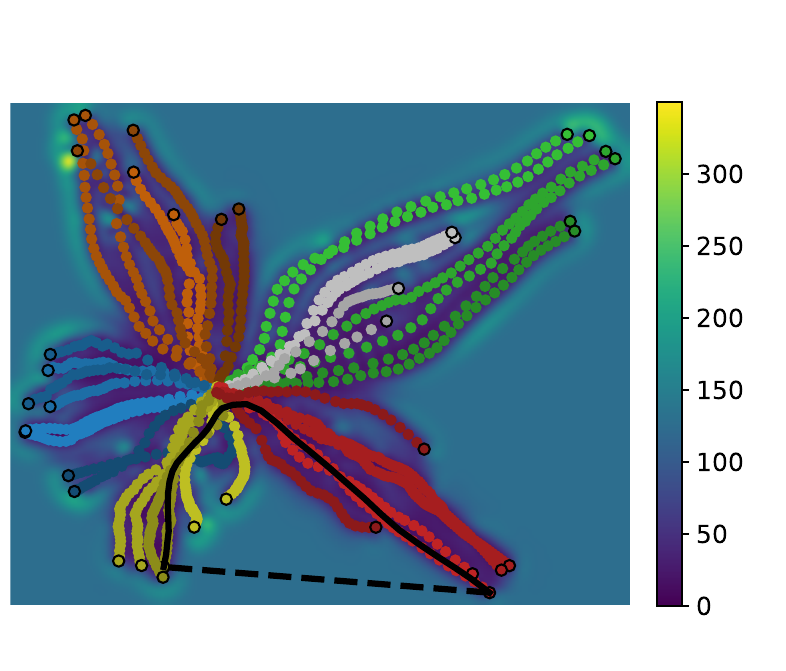}
        \verticaldashedsolidblacklinesGrasps
        \includegraphics[trim={0.1cm 0.0cm 19.5cm 12.0cm},clip,width=0.47\linewidth]{figures/hand_grasps_euclidean.pdf}
        \verticalblackdashedline
        \includegraphics[trim={0.0cm 0.2cm 0.cm 0.2cm},clip,width=0.95\linewidth]{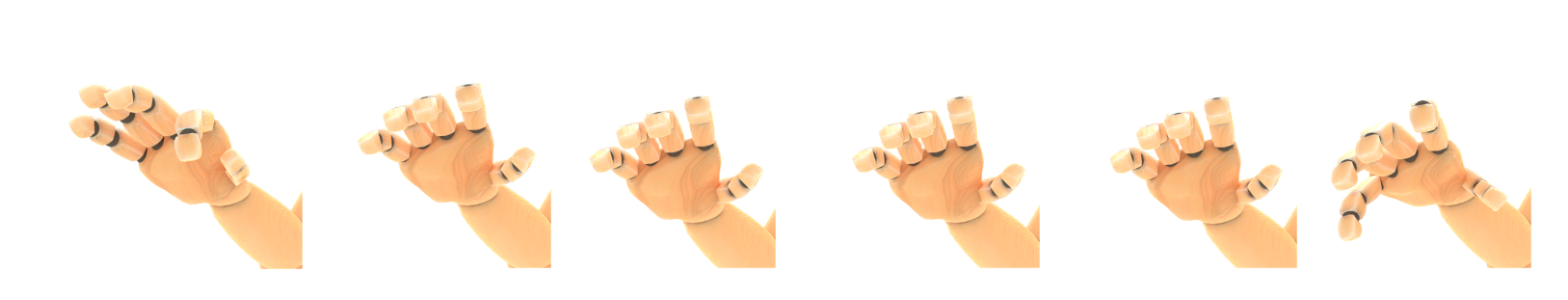}
        \verticalblackline
        \includegraphics[trim={0.0cm 0.2cm 0.cm 0.2cm},clip,width=0.95\linewidth]{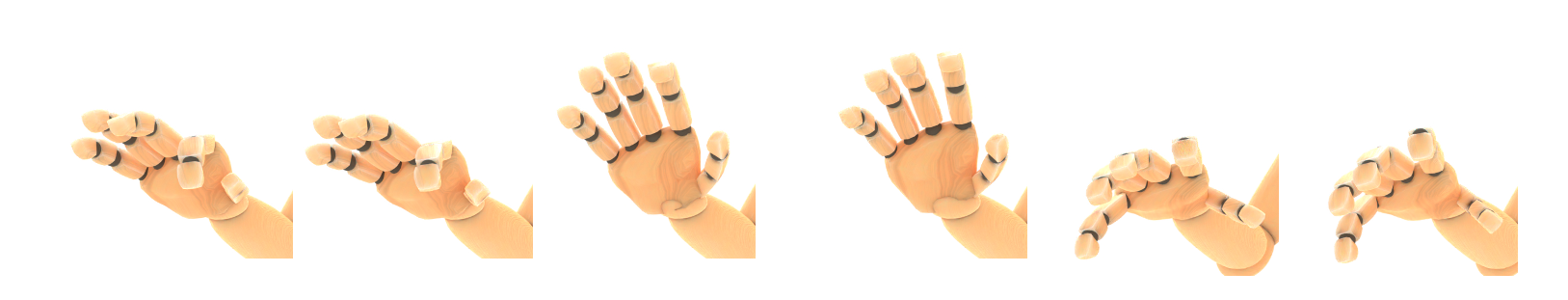}
        \caption{GPLVM, $\mathbb{R}^{2}$}
        \label{fig:hand_grasps_R2}
    \end{subfigure}
    \caption{\emph{Top left}: Embeddings of hand grasps colored according to their corresponding grasp class. The background color represents the pullback metric volume. The base manifold (\blackdashedline) and pullback (\blackline) geodesics correspond to a transition from a ring (\crimsoncircle) to a spherical grasp (\olivecircle).
    \emph{Top right}: Time-series plots of $2$ dimensions of the joint space showing the mean of the decoded geodesics with their uncertainty as a gray envelope. A training trajectory to the spherical grasp (\oliveline) and a reversed training trajectory from the ring grasp (\crimsonline) are included for reference. \emph{Bottom}: Generated hand trajectories from the decoded geodesics. }
    \vspace{-0.5cm}
    \label{fig:hand_grasps}
\end{figure*}
Jaquier et al.~\cite{Jaquier2024:GPHLVM} showed that \acp{gphlvm} outperform \acp{gplvm} to learn hand grasp embeddings that comply with the hierarchical structure of human motion taxonomies. Here, we build on this model and explore the use of pullback geodesics as a motion generation mechanism aimed at creating new motions that transition from one grasp to another.
We consider the hand grasp taxonomy of~\cite{Stival19:HumanGraspTaxonomy}, which organizes common grasps types within a hierarchical taxonomy tree. 
We use a dataset from the KIT whole-body motion database~\cite{Mandery16:KITmotionDatabase} consisting of $38$ motions of $19$ grasp types obtained from recordings of humans grasping different objects. Each motion corresponds to a subject reaching out to grasp an object from an initial resting pose and results in a trajectory $\bm{Y} \in \mathbb{R}^{N \times 24}$ representing the temporal evolution of the $24$ degrees of freedom of the wrist and fingers.\looseness=-1  

We train \acp{gphlvm} and \acp{gplvm} augmented with a taxonomy graph-distance prior~\cite{Jaquier2024:GPHLVM} and a dynamic prior akin to~\cite{Wang08:GPDM} (see App.~\ref{app-sec:hand-grasp-experiment} for details). Fig.~\ref{fig:hand_grasps} shows the resulting 2D hyperbolic and Euclidean latent spaces. Fig.~\ref{fig:hand_grasps_3D} in App.~\ref{app-sec:hand-grasp-additional-results} depicts the 3D latent spaces. All training trajectories start near the latent space origin, corresponding to the hand's initial resting pose, and progress outward until the final grasp. 
Table~\ref{tab:experiments} shows that the hyperbolic models better capture the taxonomy structure than their Euclidean counterparts, consistent with~\cite{Jaquier2024:GPHLVM}. 
Notice that the relatively high stress variance is due to the fact that the taxonomy graph distances of some grasps classes are better preserved by the embeddings than others. However, the distances are better preserved for all classes in hyperbolic latent spaces compared to Euclidean ones (see also App.~\ref{app-sec:hand-grasp-additional-results} and Fig.~\ref{fig:hand_grasps_error_matrices}).

\begin{table}[t]
    \caption{Marginal likelihood (MNIST), average stress (multi-cellular robots, hand grasps) and prediction uncertainty along decoded geodesics (all). The likelihood and the stress are compared across geometries. The prediction uncertainty is compared between the base manifold and pullback geodesic for each model.}
    \label{tab:experiments}
    \begin{center}
    \small
    \resizebox{.83\linewidth}{!}{
    \begin{tabular}{ccccc}
        \toprule
        \textbf{Experiment} & \textbf{Model} & \textbf{\textcolor{middlegreen}{Likelihood $\uparrow$}} / \textbf{Stress} $\downarrow$& \multicolumn{2}{c}{\textbf{Prediction Uncertainty ($\times 100$)} $\downarrow$} \\
        & &  & Base Manifold & Pullback \\
        \midrule
        \multirow{2}{*}{MNIST Digits} & GPLVM $\mathbb{R}^2$ & $\textcolor{middlegreen}{-226.23}$ & $9.10 \pm 4.28$ & $\underline{7.95 \pm 3.47}$ \\
        & GPHLVM $\mathbb{H}^2_\mathcal{L}$ & $\textcolor{middlegreen}{\bm{-207.14}}$ & $7.31 \pm 4.74$ &	$\bm{5.14 \pm 2.21}$\\
        \midrule
        \multirow{2}{*}{Multi-cellular Robots} & GPLVM $\mathbb{R}^2$ & $0.73 \pm 1.05$ & $3.16 \pm 2.77$ & $\underline{2.55 \pm 2.02}$ \\
        & GPHLVM $\mathbb{H}^2_\mathcal{L}$ & $\bm{0.58 \pm 0.88}$ & $4.14 \pm 4.65$ &	$\bm{0.97 \pm 1.80}$\\
        \midrule
        \multirow{4}{*}{Hand Grasps} & GPLVM $\mathbb{R}^2$ & $0.18 \pm 0.47$ & $5.24 \pm  1.99$ & $\bm{0.04 \pm 0.02}$ \\
        & GPHLVM $\mathbb{H}^2_\mathcal{L}$ & $\bm{0.10 \pm 0.26}$ & $1.50 \pm 2.27$ &	$\bm{0.04 \pm 0.02}$\\
        \cmidrule(lr){2-5}
        & GPLVM $\mathbb{R}^3$ & $0.11 \pm 0.21$ & $0.30 \pm 0.36$ & $\bm{0.01 \pm 0.01}$\\
        & GPHLVM $\mathbb{H}^3_\mathcal{L}$ & $\bm{0.07 \pm 0.10}$ & $1.29 \pm 1.50$ & $\underline{0.03 \pm 0.03}$\\
        \bottomrule
    \end{tabular}
    }
    \end{center}
    \vspace{-0.7cm}
\end{table}

Fig.~\ref{fig:hand_grasps} shows the base manifold and pullback geodesics interpolating from a ring to a spherical grasp, alongside the decoded hand motions. While the base manifold geodesic crosses empty regions in the latent space, the pullback geodesic adheres closely to the data support. 
As both geodesics start and end with the same grasps, their initial and final predictions match. However, the decoded base manifold geodesic shows high uncertainty in the middle part of the motion (see Figs.~\ref{fig:hand_grasps_H2}-\ref{fig:hand_grasps_R2} \emph{top right}). This is due to the GP reverting to its non-informative mean in regions with sparse data, where the GP prior dominates the posterior, leading to maximum uncertainty. Consequently, the obtained hand motions display large deviations from the initial ring and final spherical grasps. In contrast, as shown in Fig.~\ref{fig:hand_grasps} and Table~\ref{tab:experiments}, the decoded pullback geodesic shows a much lower uncertainty, as the geodesic stays closer to the data support. 
Moreover, the hand motions obtained from the decoded pullback geodesics produce more realistic transitions than those from the base manifold geodesics, with the hyperbolic pullback geodesics arguably exhibiting the smallest deviations from the start and end grasps. Similar results are obtained with 3D latent spaces, see App.~\ref{app-sec:hand-grasp-additional-results}.
Despite the low uncertainty, the motions predicted from 2D latent spaces sometimes lack smoothness due to the fact that nearby latent points do not always correspond to equally similar hand configurations. This is alleviated in the 3D latent space, which provides more volume to better accommodate the latent embeddings.
\looseness=-1

\vspace{-0.2cm}
\section{Conclusion}
\label{sec:conclusions}
\vspace{-0.2cm}
This paper advances the field of hyperbolic \acp{lvm} by augmenting the hyperbolic manifold with a Riemannian pullback metric, combining in a principled manner the hyperbolic geometry with the geometry of the data distribution. By minimizing the curve energy on the pullback metric, we computed geodesics that adhere to the data manifold in the hyperbolic latent space. To do so, we addressed the limitations of current auto-differentiation techniques by providing analytical solutions for the hyperbolic SE kernel derivatives. Via multiple experiments, we demonstrated that hyperbolic pullback geodesics outperform both hyperbolic geodesics and Euclidean pullback geodesics. 

It is worth noting that the benefits of pullback geodesics are most evident when the data exhibits smooth transitions. They become less effective when the data is inherently comprised of distinct clusters, as the pullback geodesics cannot cross high-energy regions, i.e., the data manifold boundaries. 
This paper focused on 2D and 3D hyperbolic latent spaces as they are easier to visualize and computationally less expensive than higher-dimensional latent spaces, while benefiting from the increased capacity of hyperbolic spaces to embed hierarchical data. Future work will investigate extensions to higher-dimensional hyperbolic latent spaces, which require to deal with more complex expressions for hyperbolic kernels.
Additionally, the need for computing manual kernel derivatives is unsatisfactory. Future work will explore alternative autodifferentiation techniques built on KeOps~\cite{Feydy20:FastSymbolicMatrices} to overcome this practical issue. Finally, the approximation of the 2D hyperbolic SE kernel is computationally expensive. Performance could potentially be increased by exploring different sampling strategies, e.g., by sampling from a Rayleigh distribution rather than from a Gaussian.\looseness-1

\clearpage
\bibliography{References}
\bibliographystyle{plain}

%%%%%%%%%%%%%%%%%%%%%%%%%%%%%%%%%%%%%%%%%%%%%%%%%%%%%%%%%%%%
\clearpage
\appendix

\section{Hyperbolic Manifold}
\label{app-sec:hyperbolic-manifold}
\begin{figure}
	\centering
	\begin{subfigure}[b]{0.34\textwidth}
		\centering
		\includegraphics[trim={0.0cm 0.0cm 0.0cm 0.0cm},clip, width=.98\textwidth]{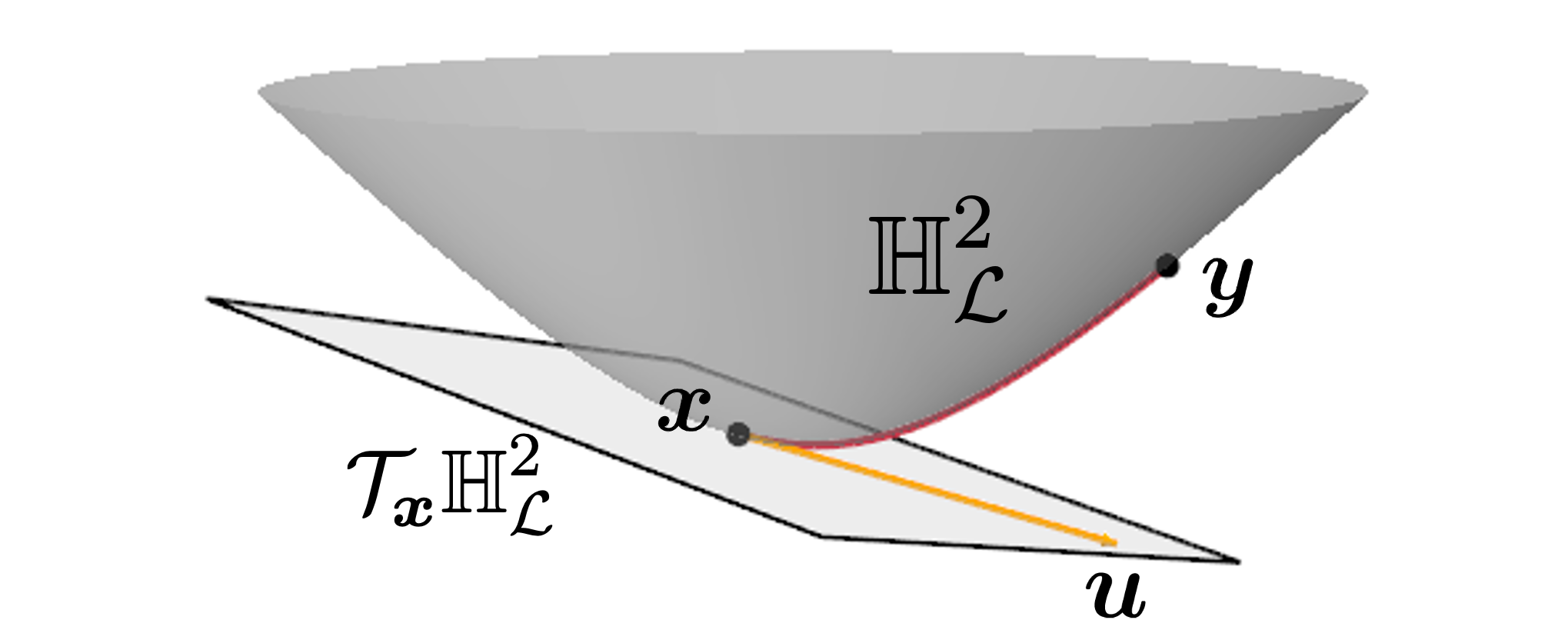}
		\caption{Exponential and logarithmic maps.}
		\label{fig:appendix:explog_maps}
	\end{subfigure}%
	\begin{subfigure}[b]{0.32\textwidth}
		\centering
		\includegraphics[trim={0.0cm 0.0cm 0.0cm 0.0cm},clip, width=\textwidth]{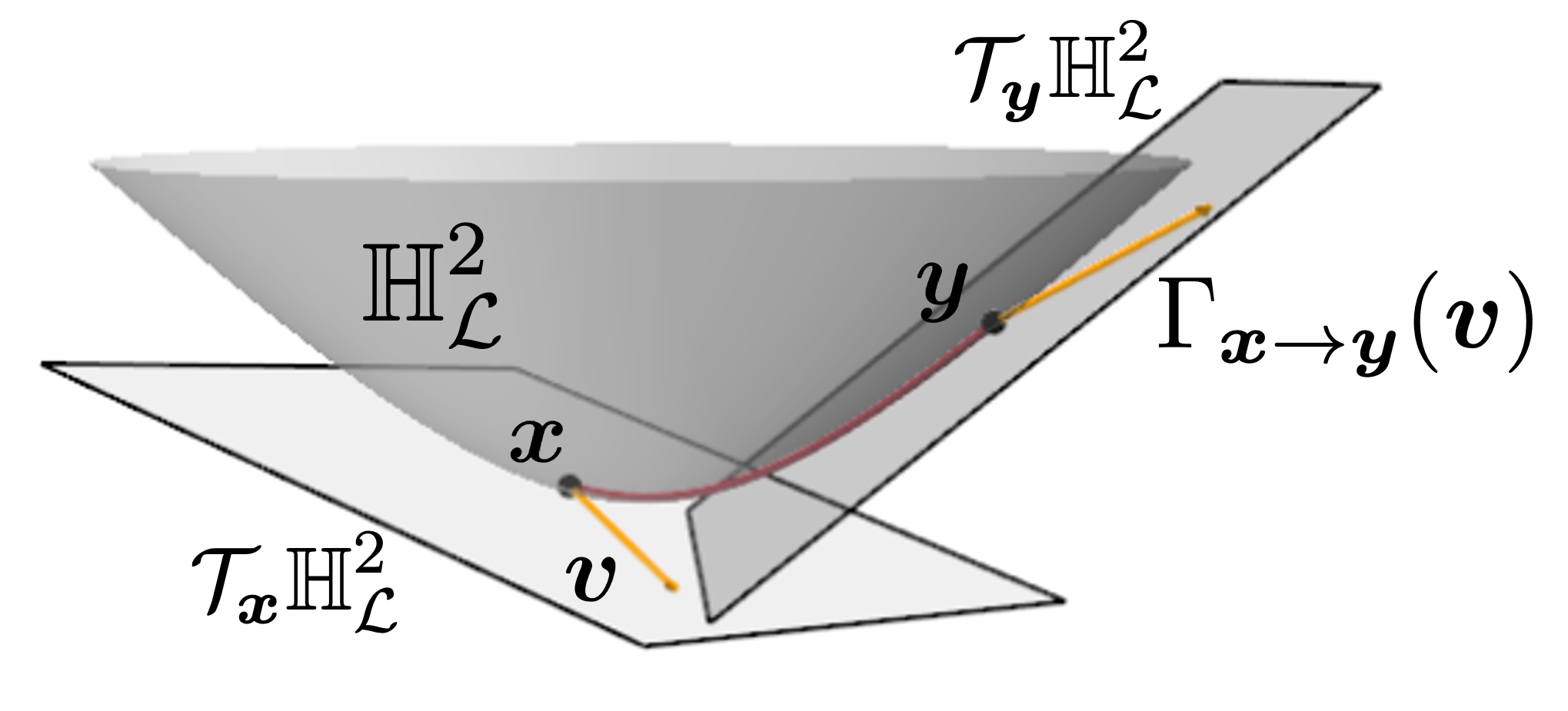}
		\caption{Parallel transport.}
		\label{fig:appendix:prl_trsp}
	\end{subfigure}
    \begin{subfigure}[b]{0.32\textwidth}
		\centering
		\includegraphics[trim={0.0cm 0.0cm 0.0cm 0.0cm},clip, width=.97\textwidth]{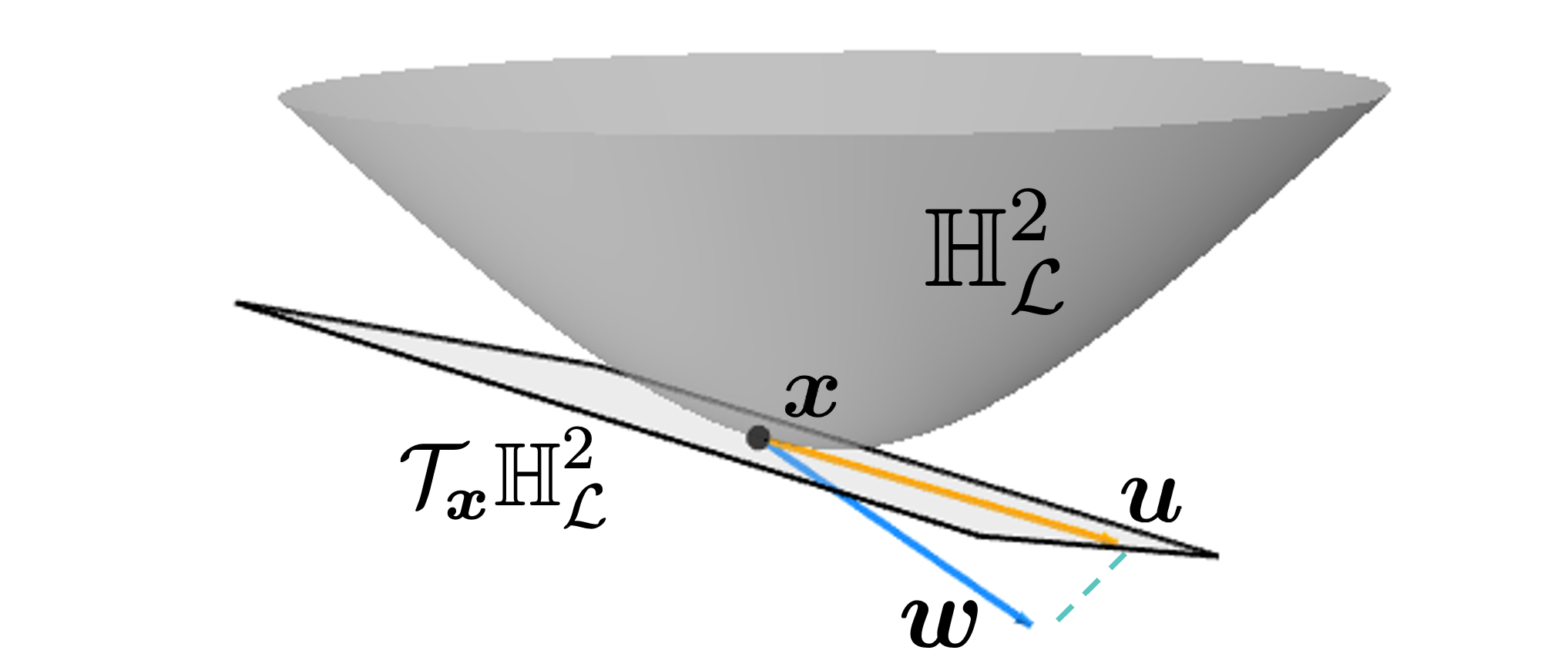}
		\caption{Tangent space projection.}
		\label{fig:appendix:proj}
	\end{subfigure}
	\caption{Principal Riemannian operations on the Lorentz model $\mathbb{H}^2_{\mathcal{L}}$. \textbf{(a)} The geodesic (\crimsonline) is the shortest path between the two points $\bm{x}$ to $\bm{y}$ on the manifold. Its length is equal to the geodesic distance $\text{d}_{\mathbb{H}^D_{\mathcal{L}}}(\bm{x}, \bm{y})$. The vector $\bm{u}$ (\yellowarrow) lies on the tangent space of $\bm{x}$ such that $\bm{y} = \expmap{\bm{x}}{\bm{u}}$. 
    \textbf{(b)} Parallel transport $\Gamma_{\bm{x} \rightarrow \bm{y}}(\bm{u})$ of the vector $\bm{v}$ from $\mathcal{T}_{\bm{x}}\mathbb{H}^2_{\mathcal{L}}$ to $\mathcal{T}_{\bm{y}}\mathbb{H}^2_{\mathcal{L}}$. \textbf{(c)} The vector $\bm{w}$ (\dodgerbluearrow) is projected onto the tangent space of $\bm{x}$ via the tangent space projection $\bm{u}=\proj_{\bm{x}} (\bm{w})$.}
	\label{fig:appendix:hyperbolic-basics}
    \vspace{-0.2cm}
\end{figure}

The $D$-dimensional hyperbolic manifold $\mathbb{H}^D_{\mathcal{L}}$ and its tangent space $\mathcal{T}_{\bm{x}} \mathbb{H}^{D}_{\mathcal{L}}$ are given as
\begin{align}
    \mathbb{H}^{D}_{\mathcal{L}} &= \{ \bm{x} \in \mathbb{R}^{D+1} \mid \langle \bm{x}, \bm{x} \rangle_\mathcal{L} = -1, x_0 > 0 \} \, , \\
    \mathcal{T}_{\bm{x}}\mathbb{H}^{D}_{\mathcal{L}} &= \{ \bm{u} \in \mathbb{R}^{D+1} \mid \langle \bm{u}, \bm{x} \rangle_{\mathcal{L}} = 0 \} \, ,
\end{align}
where $\langle \bm{x}, \bm{y} \rangle_\mathcal{L} = \bm{x}^\trsp \lorentzmetric \bm{y}$ is the Lorentzian inner product with the Lorentzian metric tensor $\lorentzmetric = \text{diag}(-1, 1, ..., 1)$. Note that, since the curvature of the hyperbolic manifold is constant, the metric tensor does not depend on $\bm{x}$ as for general manifolds, i.e., $\lorentzmetric_{\bm{x}}=\lorentzmetric$ for all $\bm{x}\in\mathbb{H}^D_{\mathcal{L}}$. For the hyperbolic manifold, closed form solutions exist for the standard manifold operations. Specifically, the geodesic distance, exponential map, logarithmic map, parallel transport, and tangent space projection are given as,
\begin{align}
    \text{d}_{\mathbb{H}^D_{\mathcal{L}}}(\bm{x}, \bm{y}) &= \text{arccosh}(-\langle \bm{x}, \bm{y} \rangle_{\mathcal{L}}) \, , \\
    \text{Exp}_{\bm{x}}(\bm{u}) &= \cosh(\Vert \bm{u} \Vert_{\mathcal{L}}) \bm{x} + \sinh(\Vert \bm{u} \Vert_{\mathcal{L}}) \dfrac{\bm{u}}{\Vert \bm{u} \Vert_{\mathcal{L}}} \, , \\
    \text{Log}_{\bm{x}}(\bm{y}) &= \text{d}_{\mathbb{H}^D_{\mathcal{L}}}(\bm{x}, \bm{y}) \dfrac{\bm{y} + \langle \bm{x}, \bm{y} \rangle_{\mathcal{L}} \bm{x}}{\sqrt{\langle \bm{x}, \bm{y} \rangle_{\mathcal{L}}^2-1}} \, , \\
    \Gamma_{\bm{x} \rightarrow \bm{y}}(\bm{u}) &= \bm{u} + \dfrac{\langle \bm{y}, \bm{u}\rangle_{\mathcal{L}}}{1 - \langle \bm{x}, \bm{y}\rangle_{\mathcal{L}}} (\bm{x} + \bm{y}) \\
    \proj_{\bm{x}} (\bm{w}) &= \bm{P}_{\bm{x}} \bm{w} = (\lorentzmetric + \bm{x}\bm{x}^\trsp)\bm{w} \, .
\end{align}
These operations are illustrated by Fig.~\ref{fig:appendix:hyperbolic-basics}.

Working with probabilistic models on Riemannian manifolds requires probability distributions that account for their geometry. Therefore, the \ac{gphlvm} prior $\bm{x}_n \sim \mathcal{N}_{\mathbb{H}^{D_x}_{\mathcal{L}}}(\bm{\mu}_0, \alpha \bm{I})$ relies on the hyperbolic wrapped distribution ~\cite{Nagano19:HyperbolicNormal}, which builds on a Gaussian distribution on the tangent space at the origin $\bm{\mu}_0 = [1 \;0 \; \ldots \: 0 ]^\trsp \in \mathbb{H}^{D}_{\mathcal{L}}$. Intuitively, the wrapped distribution is constructed as follows: \emph{(1)} Sample a point $\tilde{\bm{v}} \in \mathbb{R}^D$ from the Euclidean Gaussian distribution $\mathcal{N}(\bm{0}, \bm{\Sigma})$; \emph{(2)} Transform $\tilde{\bm{v}}$ to an element of the tangent space $\mathcal{T}_{\bm{\mu}_0}\mathbb{H}^D_{\mathcal{L}}$ at the origin by setting $\bm{v} = [ 0 \; \tilde{\bm{v}} ]^\trsp$; \emph{(3)} Parallel transport $\bm{v}$ to the desired mean $\bm{u} = \Gamma_{\bm{\mu}_0 \rightarrow \bm{\mu}}(\bm{v})$; and \emph{(4)} Project $\bm{u}$ onto the hyperbolic space via the exponential map $\bm{x} = \text{Exp}_{\bm{\mu}}(\bm{u})$. The resulting probability density function is
\begin{align}
    \label{eq:wrapped-gaussian-distribution}
    \mathcal{N}_{\mathbb{H}^D_{\mathcal{L}}}(\bm{x} \mid \bm{\mu}, \bm{\Sigma}) = \mathcal{N}(\tilde{\bm{v}} \mid \bm{0}, \bm{\Sigma}) \left( \dfrac{r}{\text{sinh}(r)} \right)^{D-1} \, ,
\end{align}
where $\bm{u} = \text{Log}_{\bm{\mu}}(\bm{x}), \bm{v} = \Gamma_{\bm{\mu} \rightarrow \bm{\mu}_0}(\bm{u})$, and $r = \Vert \bm{u} \Vert_{\mathcal{L}}$.

As mentioned in the main text, we mostly rely on the Lorentz model $\hyperbolic{D}_{\mathcal{L}}$ in our algorithms, but use the Poincaré model for visualization, as it provides a more intuitive representation. The Poincaré model is defined as
\begin{equation}
    \hyperbolic{D}_{\mathcal{P}} = \{ \bm{x}_{\mathcal{P}}\in\mathbb{R}^D \: | \: \| \bm{x}_{\mathcal{P}} \| < 1\}.
\end{equation}
It is possible to map points from the Lorentz to the Poincaré model via an isometric mapping $f:\hyperbolic{D}_{\mathcal{L}} \to \hyperbolic{D}_{\mathcal{P}}$ such that
\begin{equation}
f(\bm{x}) = \frac{\left(x_1, \dots, x_d \right)^\trsp}{x_0 + 1} ,
\label{eq:HypeDiffeomorphism}
\end{equation}
where $\bm{x} \in \hyperbolic{D}_{\mathcal{L}}$ with components $x_0, x_1, \dots, x_d$.
The inverse mapping $f^{-1}:\hyperbolic{D}_{\mathcal{P}} \to \hyperbolic{D}_{\mathcal{L}}$ is defined as follows 
\begin{equation}
f^{-1}(\bm{y}) = \frac{\left(1+\norm{}{\bm{y}}^2, 2y_1, \dots, 2y_d\right)^\trsp}{1 - \norm{}{\bm{y}^2}} ,
\end{equation}
with $\bm{y} \in \hyperbolic{D}_{\mathcal{P}}$ with components $y_1, \dots, y_d$. 

\section{Hyperbolic Kernel Derivatives}
\label{app-sec:hyperbolic-kernel-derivatives}
This section provides the complete derivations of the hyperbolic SE kernel derivatives that are necessary to compute the pullback metric tensor and to optimize geodesics on it. Eq.~\eqref{eq:gphlvm-expected-pullback-metric} defines the pullback metric tensor $\lorentzpullbackmetric_{\bm{x}^*}$ based on the Jacobian mean $\bm{\mu}_J$ and covariance matrix $\bm{\Sigma}_J$~\eqref{eq:GPHLVM-jacobian}, which in turn require the kernel derivatives $\partial k(\bm{x}^*, \bm{X}) = \frac{\partial}{\partial \bm{x}^*} k(\bm{x}^*, \bm{X}) \in \mathbb{R}^{D_x \times N}$ and $\partial^2 k(\bm{x}^*, \bm{x}^*) = \frac{\partial^2}{\partial \bm{z} \partial \bm{x}} k(\bm{x}, \bm{z})\vert_{\bm{x} = \bm{z} = \bm{x}^*} \in \mathbb{R}^{D_x \times D_x}$.
Additionally, minimizing the curve energy~\eqref{eq:curve_energy}, requires the derivative of the pullback metric tensor, i.e., the derivative of $\bm{\mu}_J^\trsp\bm{\mu}_J + D_y \bm{\Sigma}_J$, which requires the kernel derivative $\dfrac{\partial^3}{\partial \bm{x} \partial \bm{z} \partial \bm{x}} k(\bm{x}, \bm{z})$ and $\dfrac{\partial^3}{\partial \bm{z}^2 \partial \bm{x}} k(\bm{x}, \bm{z})$. Specifically, we have 
\begin{equation}
\dfrac{\partial}{\partial \bm{x}^*} ( \bm{\mu}_J^\trsp\bm{\mu}_J + D_y \bm{\Sigma}_J) = \left[ \dfrac{\partial}{\partial \bm{x}^*} \bm{\mu}_J^\trsp \right] \times_2 \bm{\mu}_J^\trsp + \left[ \dfrac{\partial}{\partial \bm{x}^*} \bm{\mu}_J \right] \times_1 \bm{\mu}_J^\trsp + D_y \left[ \dfrac{\partial}{\partial \bm{x}^*} \bm{\Sigma}_J \right], 
\end{equation}
\begin{align}
\text{with} \quad &\dfrac{\partial}{\partial \bm{x}^*} \bm{\mu}_J^\trsp = \left[ \dfrac{\partial}{\partial \bm{x}^*} \partial k(\bm{x}^*, \bm{X}) \right] \times_2 \bm{Y}^\trsp (\bm{K}_{X} + \sigma_y^2 \bm{I}_N)^{-1}, \\
&\dfrac{\partial}{\partial \bm{x}^*} \bm{\Sigma}_J = \left[ \dfrac{\partial}{\partial \bm{x}^*} \partial^2 k(\bm{x}^*, \bm{x}^*) \right] - \left[ \dfrac{\partial}{\partial \bm{x}^*} \partial k(\bm{x}^*, \bm{X}) \right] \times_2 \bm{S}_J - \left[ \dfrac{\partial}{\partial \bm{x}^*} \partial k(\bm{X}, \bm{x}^*) \right] \times_1 \bm{S}_J, \\
&\dfrac{\partial}{\partial \bm{x}^*} \partial^2 k(\bm{x}^*, \bm{x}^*) = \left[ \dfrac{\partial^3}{\partial \bm{x} \partial \bm{z} \partial \bm{x}} k(\bm{x}, \bm{z}) + \dfrac{\partial^3}{\partial \bm{z}^2 \partial \bm{x}} k(\bm{x}, \bm{z}) \right]\Bigg\vert_{\bm{x}=\bm{z}=\bm{x}^*},
\end{align}
where we have $\bm{\mu}_J \in \mathbb{R}^{D_y \times D_x}$,
$\dfrac{\partial}{\partial \bm{x}^*} \bm{\mu} \in \mathbb{R}^{D_y \times D_x \times D_x}$, $\dfrac{\partial}{\partial \bm{x}^*} \partial k(\bm{x}^*, \bm{X}) \in \mathbb{R}^{D_x \times N \times D_x}$, and $\times_n$ denotes the $n$-th mode tensor product~\cite{Kolda09:Tensor}. Next, we provide the details of the derivation of the kernel derivatives for $\mathbb{H}^{2}_{\mathcal{L}}$ and $\mathbb{H}^{3}_{\mathcal{L}}$.

\subsection{2D Hyperbolic SE Kernel Derivatives}
\label{app-subsec:2Dhyperbolic-kernel-derivatives}
We compute the 2D hyperbolic SE kernel $k^{\mathbb{H}^2_{\mathcal{L}}}(\bm{x}, \bm{z})$ via the Monte Carlo approximation~\eqref{eq:HypeKernelMonteCarlo}. The pullback metric tensor $\lorentzpullbackmetric_{\bm{x}^*}$ is computed using the kernel derivatives $\frac{\partial}{\partial \bm{x}} k(\bm{x}, \bm{z}) \vert_{\bm{x} = \bm{x}^*}$ and $\frac{\partial^2}{\partial \bm{z} \partial \bm{x}} k(\bm{x}, \bm{z}) \vert_{\bm{x}=\bm{z}=\bm{x}^*}$~\eqref{eq:kernel-derivative}. The only part of the kernel~\eqref{eq:HypeKernelMonteCarlo} that depends on the inputs $\bm{x},\bm{z}$ is the function $\Phi_l(\bm{x}, \bm{z})$. Therefore, the derivatives of the 2D hyperbolic SE kernel~\eqref{eq:HypeKernelMonteCarlo} are entirely determined by the derivatives of $\Phi_l(\bm{x}, \bm{z})$, which are given by,
\begin{equation}
   \dfrac{\partial \Phi_l(\bm{x}, \bm{z})}{\partial \bm{x}} = \left[ \dfrac{\partial \phi_l(\bm{x}_{\mathcal{P}})}{\partial \bm{x}}  \right] \bar{\phi}_l(\bm{z}_{\mathcal{P}}) \,  
   \quad \text{and} \quad
   \dfrac{\partial^2 \Phi_l(\bm{x}, \bm{z})}{\partial \bm{z} \partial \bm{x}} = \left[ \dfrac{\partial \phi_l(\bm{x}_{\mathcal{P}})}{\partial \bm{x}}  \right] \left[ \dfrac{\partial \bar{\phi}_l(\bm{z}_{\mathcal{P}})}{\partial \bm{z}}  \right]^\trsp \, .
   \label{eq:diff_x_zx_Phi}
\end{equation}
Eq~\eqref{eq:diff_x_zx_Phi} shows that these derivatives in turn are defined by the first and second derivatives of $\phi_l(\bm{x}_{\mathcal{P}}) = e^{(1+2s_li)\langle  \bm{x}_{\mathcal{P}}, \bm{b}_l\rangle}$, which are computed as
\begin{align}
\dfrac{\partial \phi_l(\bm{x}_{\mathcal{P}})}{\partial \bm{x}}  &= (1+2s_li) \,\phi_l(\bm{x}_{\mathcal{P}}) \left[ \dfrac{\partial}{\partial \bm{x}} \bm{x}_{\mathcal{P}} \right]^\trsp \left[\dfrac{\partial}{\partial \bm{x}_{\mathcal{P}}} \langle \bm{x}_{\mathcal{P}}, \bm{b} \rangle \right], \\
\dfrac{\partial^2 \phi_l(\bm{x}_{\mathcal{P}})}{\partial \bm{x}^2} 
&= (1+2s_li) \Bigg(\left[ \dfrac{\partial}{\partial \bm{x}} \bm{x}_{\mathcal{P}} \right]^\trsp \left[\dfrac{\partial}{\partial \bm{x}_{\mathcal{P}}} \langle \bm{x}_{\mathcal{P}}, \bm{b} \rangle \right] \left[ \dfrac{\partial}{\partial \bm{x}}\phi_l(\bm{x}_{\mathcal{P}}) \right]^\trsp \\
\nonumber
&\qquad \qquad \qquad +  \phi_l(\bm{x}_{\mathcal{P}}) \left[ \dfrac{\partial^2}{\partial \bm{x}^2} \bm{x}_{\mathcal{P}} \right] \times_1 \left[\dfrac{\partial}{\partial \bm{x}_{\mathcal{P}}} \langle \bm{x}_{\mathcal{P}}, \bm{b} \rangle \right] \\ 
\nonumber
&\qquad \qquad \qquad + \phi_l(\bm{x}_{\mathcal{P}}) \left[ \dfrac{\partial}{\partial \bm{x}} \bm{x}_{\mathcal{P}} \right]^\trsp \left[\dfrac{\partial}{\partial \bm{x}} \dfrac{\partial}{\partial \bm{x}_{\mathcal{P}}} \langle \bm{x}_{\mathcal{P}}, \bm{b} \rangle \right] \Bigg) , \\
&= (1+2s_li) \, \phi_l(\bm{x}_{\mathcal{P}}) \Bigg((1+2s_li)\left[ \dfrac{\partial}{\partial \bm{x}} \bm{x}_{\mathcal{P}} \right]^\trsp \left[\dfrac{\partial}{\partial \bm{x}_{\mathcal{P}}} \langle \bm{x}_{\mathcal{P}}, \bm{b} \rangle \right] \left[\dfrac{\partial}{\partial \bm{x}_{\mathcal{P}}} \langle \bm{x}_{\mathcal{P}}, \bm{b} \rangle \right]^\trsp \left[ \dfrac{\partial}{\partial \bm{x}} \bm{x}_{\mathcal{P}} \right] \\
\nonumber
&\qquad \qquad \qquad +   \left[ \dfrac{\partial^2}{\partial \bm{x}^2} \bm{x}_{\mathcal{P}} \right] \times_1 \left[\dfrac{\partial}{\partial \bm{x}_{\mathcal{P}}} \langle \bm{x}_{\mathcal{P}}, \bm{b} \rangle \right] \\ 
\nonumber
&\qquad \qquad \qquad + \left[ \dfrac{\partial}{\partial \bm{x}} \bm{x}_{\mathcal{P}} \right]^\trsp \left[\dfrac{\partial^2}{\partial \bm{x}^2_{\mathcal{P}}} \langle \bm{x}_{\mathcal{P}}, \bm{b} \rangle \right] \left[ \dfrac{\partial}{\partial \bm{x}}\bm{x}_{\mathcal{P}} \right] \Bigg) \, .
\end{align}
As shown by these expressions, the derivatives of $\phi_l(\bm{x}_{\mathcal{P}})$ are defined as functions of the derivatives of the hyperbolic outer product and of the derivatives of the Lorentz to Poincaré mapping. The derivatives of the hyperbolic outer product $\langle \bm{x}_\mathcal{P}, \bm{b} \rangle = \frac{1}{2} \text{log}\left( \frac{1 - \vert \bm{x}_{\mathcal{P}}\vert^2}{\vert \bm{x}_{\mathcal{P}} - \bm{b} \vert^2} \right)$ are given as, 
\begin{align}
\dfrac{\partial}{\partial \bm{x}_{\mathcal{P}}} \langle \bm{x}_{\mathcal{P}}, \bm{b} \rangle 
&= -\dfrac{1}{2} \left(\dfrac{2\bm{x}_{\mathcal{P}}}{1-\vert \bm{x}_{\mathcal{P}}\vert^2} + \dfrac{2 (\bm{x}_{\mathcal{P}}-\bm{b})}{\vert \bm{x}_{\mathcal{P}} - \bm{b}\vert^2}\right) = \dfrac{\bm{x}_{\mathcal{P}}}{\vert \bm{x}_{\mathcal{P}} \vert^2 -  1} - \dfrac{ \bm{x}_{\mathcal{P}} - \bm{b}}{\vert \bm{x}_{\mathcal{P}} - \bm{b} \vert^2}, \\
\dfrac{\partial^2}{\partial \bm{x}_{\mathcal{P}}^2} \langle \bm{x}_{\mathcal{P}}, \bm{b} \rangle 
&= \dfrac{1}{\vert \bm{x}_{\mathcal{P}} \vert^2-1} \bm{I}_{D_x} - \dfrac{2\bm{x}_{\mathcal{P}}\bm{x}_{\mathcal{P}}^\trsp}{(\vert \bm{x}_{\mathcal{P}} \vert^2-1)^2} - \dfrac{1}{\vert \bm{x}_{\mathcal{P}} - \bm{b} \vert^2} \bm{I}_{D_x} + \dfrac{2(\bm{x}_{\mathcal{P}} - \bm{b})(\bm{x}_{\mathcal{P}}-\bm{b})^\trsp}{\vert \bm{x}_{\mathcal{P}} - \bm{b} \vert^4 } \, .
\end{align}
Finally, the derivatives of the Lorentz to Poincaré mapping $\bm{x}_{\mathcal{P}} \!= \!\frac{1}{1 + x_0} [x_1 \ldots x_D ]^\trsp \in \mathbb{H}^{D_x}_{\mathcal{P}}$ are given as,
\begin{align}
\left(\dfrac{\partial}{\partial \bm{x}} \bm{x}_{\mathcal{P}}\right)_{i,j} &= \begin{cases}
-\frac{x_{i+1}}{(1+x_0)^2} & j=0 \\
\quad \frac{1}{1+x_0} & j=i+1 \\
\quad 0 & \text{otherwise}
\end{cases} \quad , \quad
\left(\dfrac{\partial^2}{\partial \bm{x}^2} \bm{x}_{\mathcal{P}}\right)_{i,j,k} = \begin{cases}
\quad\frac{2x_{i+1}}{(1+x_0)^3} & j=k=0 \\
- \frac{1}{(1+x_0)^2} & j=0 \text{ and } k=i+1 \\
- \frac{1}{(1+x_0)^2} & k=0 \text{ and } j = i+1\\
\quad 0 & \text{otherwise}
\end{cases}  \, ,
\end{align}
with $i \in \{0, ..., D_x-1 \}$, $ j,k \in \{0, ..., D_x \}$, and $D_x=2$.

Minimizing the curve energy~\eqref{eq:curve_energy} to compute pullback geodesics additionally requires the derivative $\frac{\partial}{\partial \bm{x}^*} \lorentzpullbackmetric_{ \bm{x}^*}$. This, in turn, requires the derivative of the Jacobian of the covariance matrix $\frac{\partial \bm{\Sigma}_J}{\partial \bm{x}^*}$, which depends on the kernel derivatives $\frac{\partial^3 k(\bm{x}, \bm{z})}{\partial \bm{x} \partial \bm{z} \partial \bm{x}} \vert_{\bm{x}=\bm{z}=\bm{x}^*}$, and $\frac{\partial^3 k(\bm{x}, \bm{z})}{\partial \bm{z}^2 \partial \bm{x}}  \vert_{\bm{x}=\bm{z}=\bm{x}^*}$. Again, the kernel derivatives are determined by the derivatives of $\Phi_l(\bm{x}, \bm{z})$, 
\begin{equation}
    \frac{\partial^3 \Phi_l(\bm{x}, \bm{z})}{\partial \bm{x} \partial \bm{z} \partial \bm{x}} = \left[ \dfrac{\partial^2 \phi_l(\bm{x}_{\mathcal{P}} )}{\partial \bm{x}^2}  \right] \times_2 \left[ \dfrac{\partial \bar{\phi}_l(\bm{z}_{\mathcal{P}} )}{\partial \bm{z}} \right] 
    \quad \text{and} \quad
    \frac{\partial^3 \Phi_l(\bm{x}, \bm{z})}{\partial \bm{z}^2 \partial \bm{x}} = \left[ \dfrac{\partial^2 \bar{\phi}_l(\bm{z}_{\mathcal{P}})}{\partial \bm{z}^2}  \right] \times_1 \left[ \dfrac{\partial \phi_l(\bm{x}_{\mathcal{P}})}{\partial \bm{x}}  \right] ,
    \label{eq:diff_xzx_zzx_Phi}
\end{equation}
where $\bm{A} \times_i \bm{B}$ refers to n-mode product~\cite{Kolda09:TensorDecompositions} which multiplies the $i$-th dimension of $\bm{A}$ with the second dimension of $\bm{B}$. If $\bm{B}$ is a vector, $\bm{A}$ is unsqueezed at dimension $i$ before the multiplication. The matrices $\frac{\partial^2 \phi_l(\bm{x}_\mathcal{P})}{\partial \bm{x}^2} , \frac{\partial^2 \phi_l(\bm{z}_\mathcal{P})}{\partial \bm{z}^2}  \in \mathbb{R}^{3 \times 3}$ in~\eqref{eq:diff_xzx_zzx_Phi} are therefore interpreted as $\mathbb{R}^{3 \times 1 \times 3}$ and $\mathbb{R}^{1 \times 3 \times 3}$, respectively.

\subsection{3D Hyperbolic SE Kernel Derivatives}
\label{app-subsec:3Dhyperbolic-kernel-derivatives}
This section provides the derivations of the 3D hyperbolic SE kernel derivatives. The 3D hyperbolic SE kernel is given by~\eqref{eq:heat-kernel-2-3D}. In this section, we simplify its notation and write $k(\bm{x}, \bm{z}) = \frac{\rho}{s} e^{-\frac{\rho^2}{\nu}}$ by omitting the constant $\tau/C_\infty$, setting $\nu=2\kappa^2$, and introducing the helper variables $u = \langle \bm{x}, \bm{z} \rangle_{\mathcal{L}}$ and $s = \sqrt{u^2-1}$. Their derivatives are $\frac{\partial u}{\partial \bm{x}} = \lorentzmetric\bm{z}$, $\frac{\text{d} \rho}{\text{d} u} = \frac{-1}{s}$, and $\frac{\text{d} s}{\text{d} u} = \frac{u}{s}$. Using these expressions, the first and second derivatives of the 3D hyperbolic SE kernel $k(\bm{x}, \bm{y}) = \frac{\rho}{s} \, e^{-\frac{\rho^2}{\nu}}$ are given as,
\begin{align}
\dfrac{\partial}{\partial \bm{x}}\, k(\bm{x}, \bm{z}) 
&= \left[\dfrac{\partial u}{\partial \bm{x}}\dfrac{\text{d}}{\text{d} u}\rho \right] \dfrac{1}{s} e^{-\frac{\rho^2}{\nu}} + \rho \left[ \dfrac{\partial u}{\partial \bm{x}} \dfrac{\text{d} }{\text{d} u} s^{-1} \right] e^{-\frac{\rho^2}{\nu}} + \dfrac{\rho}{s}  \left[ \dfrac{\partial u}{\partial  \bm{x}} \dfrac{\text{d} \rho }{\text{d} u} \dfrac{\text{d}}{\text{d} \rho} e^{-\frac{\rho^2}{\nu}} \right] , \nonumber\\
&= -\dfrac{1}{s^2} \lorentzmetric \bm{z} e^{-\frac{\rho^2}{\nu}} - \dfrac{\rho u}{s^3} \lorentzmetric \bm{z} e^{-\frac{\rho^2}{\nu}} + \dfrac{\rho}{s^2} \lorentzmetric \bm{z} \dfrac{2\rho}{\nu} e^{-\frac{\rho^2}{\nu}} , \nonumber\\
&= g(u) e^{-\frac{\rho^2}{\nu}}\lorentzmetric \bm{z} \, , \label{eq:diff_x_k_3D} \\
\dfrac{\partial^2}{\partial \bm{z} \partial{\bm{x}}} k(\bm{x}, \bm{z})
&= \bm{Gz} \left[ \dfrac{\partial u}{\partial \bm{z}} \dfrac{\text{d}}{\text{d} u} g(u) \right] e^{-\frac{\rho^2}{\nu}} + \bm{Gz}  g(u) \left[ \dfrac{\partial u}{\partial \bm{z}} \dfrac{\text{d} \rho }{\text{d} u}  \dfrac{\text{d}}{\text{d} \rho} e^{-\frac{\rho^2}{\nu}} \right] + g(u) e^{-\frac{\rho^2}{\nu}} \left[\dfrac{\partial}{\partial \bm{z}} \bm{Gz} \right] , \nonumber\\
&= \left[\dfrac{\text{d}}{\text{d} u} g(u) \right] e^{-\frac{\rho^2}{\nu}} \bm{Gz}\bm{x}^\trsp \lorentzmetric + g(u) \dfrac{-1}{s} \dfrac{-2\rho}{\nu} e^{-\frac{\rho^2}{\nu}} \bm{Gz}\bm{x}^\trsp \lorentzmetric + g(u) e^{-\frac{\rho^2}{\nu}} \lorentzmetric ,\nonumber \\
&= \left(h(u) \lorentzmetric \bm{z}\bm{x}^\trsp \lorentzmetric  + g(u) \lorentzmetric \right) e^{-\frac{\rho^2}{\nu}} \, ,  \label{eq:diff_zx_k_3D} \\
\dfrac{\partial^2}{\partial \bm{x} \partial \bm{x}} k(\bm{x}, \bm{z}) &= h(u)\lorentzmetric \bm{z}\bm{z}^\trsp \lorentzmetric e^{-\frac{\rho^2}{\nu}} \, , \label{eq:diff_xx_k_3D} 
\end{align}
with $g(u) \!=\! \left(\frac{2\rho^2}{\nu s^2} - \frac{1}{s^2} - \frac{u\rho}{s^3} \right)$ and $h(u) \!=\! \frac{\text{d}}{\text{d}u} g(u) + \frac{g(u) 2\rho}{\nu s}$. The helper functions $g(u)$ and $h(u)$ are essential to understand why standard automatic differentiation tools fail to compute the derivatives~\eqref{eq:diff_x_k_3D}-\eqref{eq:diff_xx_k_3D}. For equal inputs $\bm{x} \!=\! \bm{z}$, the inner product $u$ approaches $-1$, while the distance $\rho = \text{d}_{\mathbb{H}^{D_x}_{\mathcal{L}}}(\bm{x}, \bm{z})$ and variable $s$ converge to $0$. This limit is analytically well behaved for the helper functions, e.g., $\lim_{u \rightarrow -1^-} g(u) = \frac{2}{\nu} + \frac{1}{3}$. However, automatic differentiation tools fail to compute the kernel derivatives for equal inputs correctly because they do not cancel out $0$-approaching terms. For example, we have for $g(u)$,
\begin{equation*}
    \dfrac{2\rho^2}{\nu s^2} - \dfrac{1}{s^2} - \dfrac{u\rho}{s^3} \rightarrow^{\text{ autodiff}}_{\: \: \: \:\bm{x}=\bm{z}} \: \rightarrow \dfrac{0}{0} - \dfrac{1}{0} - \dfrac{0}{0} = \text{NaN} \, .
\end{equation*}
In this scenario, autodiff divides by $0$, leading to undefined values (NaN). In other cases, e.g., when training a \ac{gplvm}, derivatives for equal kernel inputs $\bm{x}=\bm{z}$ may not be that relevant because different latent points rarely become that close. However, these derivatives are essential for computing the Jacobian covariance matrix $\bm{\Sigma}_J$ in~\eqref{eq:GPHLVM-jacobian}. 
Although symbolic differentiation libraries could provide a solution, they tend to be too slow for practical purposes. 
To address this issue, we replace the general function definitions with their analytical limits, when $u$ gets close to $-1$. This is equivalent to the hyperbolic distance between the two kernel inputs falling below a predefined threshold value $\rho = \text{arccosh}(-u) = \text{d}_{\mathbb{H}^{D_x}_{\mathcal{L}}}(\bm{x}, \bm{z}) \le 1\mathrm{e}{-4}$. The analytical limit expressions for the helper functions are given as
\begin{alignat}{3}
\lim_{u \rightarrow -1^-} g(u) &= \lim_{u \rightarrow -1^-} \left( \frac{2\rho^2}{\nu s^2} - \frac{1}{s^2} - \frac{u\rho}{s^3} \right) &&= \frac{2}{\nu} + \dfrac{1}{3} \, , \\
\lim_{u \rightarrow -1^-} h(u) &= \lim_{u \rightarrow -1^-} \left( \dfrac{\text{d}}{\text{d}u} g(u) + \frac{g(u) 2\rho}{\nu s} \right) &&= \dfrac{4}{\nu^2} + \dfrac{6}{3\nu} + \dfrac{4}{15} \, , \\
\lim_{u \rightarrow -1^-} q(u) &= \lim_{u \rightarrow -1^-} \left( \dfrac{\text{d}}{\text{d}u} h(u) + \frac{h(u) 2\rho}{\nu s} \right) &&= \dfrac{8}{\nu^3} + \dfrac{8}{\nu^2} + \dfrac{14}{5\nu} + \dfrac{12}{35} \, .
\end{alignat}

Finally, we need the third-order kernel derivatives to optimize pullback geodesics as in the 2D case. Denoting the second kernel derivative as $\bm{K} = \frac{\partial^2 k(\bm{x}, \bm{z})}{\partial \bm{z} \partial \bm{x}} \in \mathbb{R}^{4 \times 4}$, the third-order derivatives is obtained by stacking the individual derivatives of each row and column of $\bm{K}$. The derivative of each row $\bm{K}_i$ w.r.t. the first input $\bm{x}$ and each column $\bm{K}_j$ w.r.t the second input $\bm{z}$ are given by, 
\begin{align}
\dfrac{\partial}{\partial \bm{z}}\bm{K}_j &= \dfrac{\partial}{\partial \bm{z}} \big(h(u) \lorentzmetric \bm{z}x_j + g(u) \bm{e}_j \big)\Delta_{1j} e^{-\frac{\rho^2}{\nu}}  ,\\
&= \left(\left[\dfrac{\text{d}}{\text{d}u} h(u) \right] \lorentzmetric \bm{z}\bm{x}^\trsp \lorentzmetric x_j + h(u)\lorentzmetric x_j + \left[ \dfrac{\text{d}}{\text{d}u}g(u) \right] \bm{e}_j \bm{x}^\trsp \lorentzmetric \right) \Delta_{1j} e^{-\frac{\rho^2}{\nu}} \\
&\quad +  \big(h(u) \lorentzmetric \bm{z}x_j + g(u) \bm{e}_j \big)  \Delta_{1l} \dfrac{2\rho}{\nu s}e^{-\frac{\rho^2}{\nu}} \bm{x}^\trsp \lorentzmetric ,\\
&= \left(q(u) \lorentzmetric \bm{z}\bm{x}^\trsp \lorentzmetric x_j + h(u)\lorentzmetric x_j + h(u) \bm{e}_j \bm{x}^\trsp \lorentzmetric \right) \Delta_{1j} e^{-\frac{\rho^2}{\nu}} \, ,\\
\dfrac{\partial}{\partial \bm{x}} \bm{K}_i &= \left(q(u) \lorentzmetric \bm{z}\bm{x}^\trsp \lorentzmetric z_i \, + h(u)\lorentzmetric z_i \, + h(u) \bm{Gz}\bm{e}_i^\trsp \right) \Delta_{1i} e^{-\frac{\rho^2}{\nu}} \, , 
\end{align}
where $q(u) = \frac{\text{d}}{\text{d}u} h(u) + \frac{h(u)2\rho}{\nu s}$ and $\Delta_{ij}$ represents a modified Kronecker delta equal to $-1$ for equal indices and to $1$ otherwise. Stacking the row and column matrices $\bm{K}_i$ and $\bm{K}_j$ along the first and second dimension accordingly gives the third kernel derivatives $\frac{\partial \bm{K}}{\partial \bm{x}} , \frac{\partial \bm{K}}{\partial \bm{z}} \in \mathbb{R}^{4 \times 4 \times 4}$.

\section{Hyperbolic Pullback Geodesics}
\label{app:pullback-geodesics}
In Sec.~\ref{sec:pullback-geodesics}, we described the computation of geodesics according to the hyperbolic pullback metric introduced in Sec.~\ref{sec:lvm-hyperbolic-pullback},~\ref{sec:gplvm-hyperbolic-pullback}. 
Algorithm~\ref{alg:hyperbolic-pullback-geodesics} summarizes the computation of a hyperbolic pullback geodesic by minimizing~\eqref{eq:loss_energy_regularized} with Riemannian optimization.

\begin{algorithm}[tb]
   \caption{Computation of hyperbolic pullback geodesics}
   \label{alg:hyperbolic-pullback-geodesics}
\begin{algorithmic}
   \STATE {\bfseries Input:} Start and end points $\bm{x}_1$,  $\bm{x}_M\in \mathbb{H}^{D_x}_{\mathcal{L}}$ of the pullback geodesic, pullback metric $\lorentzpullbackmetric$.
   \STATE {\bfseries Output:} Pullback geodesic represented as a set of points $\{\bm{x}_i\}_{i=1}^M$ with $\bm{x}_i \in \mathbb{H}^{D_x}_{\mathcal{L}}$.
   \STATE {\bfseries Initialization:}
   \STATE Initialize the pullback geodesic, e.g., as the hyperbolic geodesic from $\bm{x}_1$ to $\bm{x}_M$.
   \STATE Discretize the pullback geodesic with $M$ points $\{\bm{x}_i\}_{i=1}^M$.
   \STATE {\bfseries Geodesic computation:}
   \REPEAT
   \STATE Compute the curve energy $E$~\eqref{eq:curve_energy}.
   \STATE Compute the spline energy $E_{\text{spline}}$.
   \STATE ${\bm{x}_2, ..., \bm{x}_{M-1}} \gets \mathsf{RiemannianOptStep}( E + \lambda E_{\text{spline}})$.
   \UNTIL{convergence}
\end{algorithmic}
\end{algorithm}

\section{Experimental Details and Additional Results}
\label{app-sec:experimental-details}
Apps.~\ref{app-sec:Cshape}-\ref{app-sec:hand-grasp-experiment} provide additional details regarding the data and \ac{gphlvm} training process for each of the experiments described in Sec. \ref{sec:experiments}. App.~\ref{app-sec:hand-grasp-additional-results} provides additional insights into the hand grasps experiment of Sec.~\ref{sec:hand-grasps-generation}, and App.~\ref{app-subsec:runtimes} provides runtimes for the computation of pullback metrics and geodesics.

\subsection{C-shape}
\label{app-sec:Cshape}
The proof-of-concept experiment on the $\mathsf{C}$-shape dataset was designed as a minimalist example with a simple setup to visualize and compare Euclidean and hyperbolic pullback metrics. The training data consists of $N=1000$ two-dimensional points arranged in a $\mathsf{C}$ shape and scaled to fit within the unit circle. In the Euclidean case, the datapoints are used as both latent points and observations, i.e., $\bm{X} = \bm{Y} \in \mathbb{R}^{1000 \times 2}$. In the hyperbolic case, the datapoints can be interpreted as elements of the Poincaré model. We represent each latent point and observation in the Lorentz model using the Poincaré to Lorentz mapping $\bm{x}_{\mathcal{L}} = \frac{1}{1 - \Vert \bm{x} \Vert^2} \begin{bmatrix} 1 + \Vert \bm{x} \Vert^2 & 2 \bm{x}^\trsp \end{bmatrix}^\trsp \in \mathbb{H}^{2}_{\mathcal{L}}$. We fully specify the \acp{lvm} by manually setting the kernel variance, lengthscale, and noise variance of both \ac{gplvm} and \ac{gphlvm} to $(\tau, \kappa, \sigma^2_y) = (0.7, 0.15, 0.69)$. We used constant zero mean for both models. For the \ac{gphlvm}, we used the 2D hyperbolic SE kernel with $L = 3000$ rejection samples.  

The start- and end-points of the geodesics shown in Fig.~\ref{fig:c-shape} correspond to one of the $\mathsf{C}$-shape latent points. The geodesics are represented by $M=25$ regularly-spaced points. Geodesics on the Euclidean base manifold correspond to straight lines from $\bm{x}_0$ to $\bm{x}_{M-1}$. In the hyperbolic case, the geodesics on the base manifold are given as $\bm{x}_i = \text{Exp}_{\bm{x}_0}(t_i \text{Log}_{\bm{x}_0}(\bm{x}_{M-1}))$ with $t_i = i / (M-1)$ and $i=\{0, ..., M-1 \}$. The geodesics on the base manifolds were used as initialization for the pullback geodesic optimization.  
We obtain the pullback geodesics by minimizing the combined loss function~\eqref{eq:loss_energy_regularized} based on both curve energy and spline energy (see Sec.~\ref{sec:pullback-geodesics}). We used Riemannian Adam for $200$ steps with a learning rate of $0.005$. In this experiment, the curve energy and spline energy were weighted equally, i.e., $\lambda = 1$ in Eq.~\eqref{eq:curve_energy}.

We visualize the volume of the pullback metrics on a $110 \times 110$ grid of points, filtering out points lying outside the unit circle in the hyperbolic case. In the Euclidean case, we computed the volume at each grid point $\bm{x}^*$ by evaluating the determinant $\sqrt{ \text{det} (\euclpullbackmetric_{\bm{x}^*})}$ of the pullback metric tensor $\euclpullbackmetric$. In the hyperbolic case, the pullback metric tensor is a $3 \times 3$ matrix that lies on the $2$-dimensional tangent space, leading to one of its eigenvalues being zero. To address this, we compute the volume $\sqrt{ \text{det} (\lorentzpullbackmetric_{\bm{x}^*})}$ as the product of the two nonzero eigenvalues.

\subsection{MNIST Digits}
\label{app-sec:mnist}
We use a subset of the MNIST dataset composed of $100$ datapoints for each of the classes $0$, $1$, $2$, $3$, $6$, and $9$. Each datapoint is represented as a $28\times 28$ binary image. 
We optimize the parameters of the \ac{gphlvm} using Riemannian Adam for $500$ steps with a learning rate of $0.05$. We set a Gamma prior with concentration $\alpha=2$ and rate $\beta=2$ on the kernel lengthscale and a Gamma prior with concentration $\alpha=5$ and rate $\beta=0.8$ on the kernel variance. The hyperbolic embeddings are initialized in the tangent space of the origin $\bm{\mu}_0$ using PCA and projected onto the manifold using the exponential map.
Similarly, we optimize the parameters of the \ac{gplvm} using Adam for $500$ steps with a learning rate of $0.01$. The Euclidean embeddings are initialized using PCA.

\subsection{Multi--cellular Robot Design}
\label{app-sec:multicellular}
We consider multi--cellular robots composed of a $5\times 5$ grid of cells. Possible cell types are empty, solid, soft, a horizontal actuator or a vertical actuator. We generate a dataset of $216$ multi-cellular robots following the hierarchical approach of~\cite{Dong24:RobotDesign}, which we explain next for completeness. The root nodes of the hierarchy are two coarse robots composed of only vertical and only horizontal actuator cells. A component is defined as a group of cells of the same type, i.e., the root robots have a single component. Children nodes of each robot are designed by dividing large components into $2$ smaller components using K-means and by assigning a different type of cell to one of these components. This process is repeated iteratively to obtain robots with up to $8$ components. This results in a tree of multi-cellular robots, with coarse-grained robots close to the roots and fine-grained robots at the leaves.\looseness=-1

To account for the hierarchical structure associated with the tree structure of the multi-cellular robots, we incorporate graph-based priors in both \ac{gphlvm} and \ac{gplvm} as proposed in~\cite{Jaquier2024:GPHLVM}. This is achieved by leveraging the stress loss, 
\begin{equation} 
\ell_{\text{stress}}(\bm{X}) = \sum_{i < j} \left( \text{d}_{\mathbb{G}}(c_i, c_j) - \text{d}_{\mathcal{M}}(\bm{x}_i, \bm{x}_j) \right)^2,
\label{eq:stress_loss}
\end{equation}
to match the pairwise latent distances with the tree distances, and where $c_i$ denotes the tree node, $\operatorname{dist}_{\mathbb{G}}$, and $\operatorname{dist}_{\manifold}$ are the graph distance and the geodesic distance on $\manifold$, respectively.
Therefore, training the \ac{gphlvm} and \ac{gplvm} is achieved by solving,
\begin{equation}
    \argmax_{\bm{X}} \log p(\bm{Y} \mid \bm{X}, \Theta)  - \gamma \bar{\ell}_{\text{stress}}(\bm{X}) \, ,
\end{equation}
where $\gamma$ is a parameter balancing the two losses and $\bar{\ell}_{\text{stress}}(\bm{X})$ is the stress loss averaged over the embeddings. We set $\gamma=6000$. For the \ac{gphlvm}, we use a Gamma prior with concentration $\alpha=2$ and rate $\beta=2$ on the kernel lengthscale. We optimize the parameters of the \ac{gphlvm} using Riemannian Adam for $500$ steps with a learning rate of $0.05$ and the parameters of the \ac{gplvm} using Adam for $500$ steps with a learning rate of $0.01$.

\subsection{Hand Grasps Generation}
\label{app-sec:hand-grasp-experiment}
For the hand grasp experiment, we used data from the KIT Whole-Body Motion Database~\cite{Mandery16:KITmotionDatabase}. The dataset consists of $38$ trajectories, with motion recordings from two human subjects (IDs $2122$ and $2123$). Each subject performed $19$ different grasp types\footnote{The three-finger sphere grasp of the GRASP taxonomy~\cite{Feix16:GRASPtaxonomy} is missing from the dataset.} of the GRASP taxonomy~\cite{Feix16:GRASPtaxonomy}. Each grasp consists of the subject grasping an object from a table, lifting it, and placing it back. 

We applied several preprocessing steps to the recorded data before training the \ac{gphlvm}: \emph{(1)} We applied a low-pass filter to remove high-frequency noise; \emph{(2)} We trimmed the start and end of each trajectory to focus solely on the motion from the initial resting pose to the point where the grasp was completed. Since detecting the exact point of grasp completion is non-trivial, we cut off the trajectories at the moment when the grasped object was first moved by the subject; \emph{(3)} We subsample the trajectories; and \emph{(4)} We centered the data to allow for the use of a zero mean function in the \ac{gphlvm}. After preprocessing, each trajectory is composed of $30$ to $40$ data points. Stacking all trajectories together results in a total of $N=1321$ data points $\bm{Y} \in \mathbb{R}^{1321 \times 24}$. 

In contrast to the previous experiments, this experiment involves trajectory data. To preserve the trajectory structure during training, we incorporated a dynamics prior  $p(\bm{X}_{2:N} \mid \bm{X}_{1:N-1})$, similar to~\cite{Wang08:GPDM}, but using the wrapped Gaussian distribution~\eqref{eq:wrapped-gaussian-distribution}. We also used back constraints $\bm{X} = k(\bm{Y}, \bm{Y}) \bm{C}$, which allow for a smooth mapping from the observation space to the latent space. The back-constraints kernel is defined as a Euclidean SE kernel with variance $\tau=1$ and a lengthscale $\kappa=0.4$. 
Moreover, each grasp type is identified with a lead node of the quantitative taxonomy of hand grasps~\cite{Stival19:HumanGraspTaxonomy}. More specifically, the first point of each trajectory, which corresponds to the resting state, is assigned to the root node $c_0$ of the taxonomy graph, and the last point to the taxonomy node representing the corresponding grasp type. The taxonomy node $c_{14}$, for instance, represents the stick grasp. The number of edges between different grasp types in the taxonomy graph defines a distance function, $\text{d}_{\mathbb{G}}(c_i, c_j)$. 
To account for the hierarchical structure associated with the taxonomy, we additionally incorporate graph-based priors in our \ac{gphlvm} as proposed by Jaquier et al.~\cite{Jaquier2024:GPHLVM}. This is achieved by leveraging the stress loss~\eqref{eq:stress_loss} to match the pairwise latent distances with the graph distances. 
During training, we minimize the stress loss over the first and last latent points of each trajectory along with maximizing the \ac{gphlvm} marginal likelihood by optimizing,
\begin{equation}
    \argmax_{\bm{X}} \log p(\bm{Y} \mid \bm{X}, \Theta) + 2\frac{D_y}{D_x} \log p(\bm{X}_{2:N} \mid \bm{X}_{1:N-1}) - D_y\ell_{\text{stress}}(\bm{X}) \, .
\end{equation}
We performed the optimization using Riemannian Adam for $10000$ steps, with a learning rate of $0.001$.

As \acp{gplvm} are generally prone to local optima during training, they benefit from a good initialization. Therefore, we initialize the first and last latent points of each trajectory to minimize the stress loss, i.e.,\looseness=-1
\begin{align} 
\bm{X}_{\text{init}} = \min_{\bm{X}} \ell_{\text{stress}}(\bm{X}). 
\end{align}
The optimized first and last points are then connected using a hyperbolic geodesic with the same number of points as the original motion recording. To ensure distinct initialization for each subject, we added random noise to the geodesics.

For the geodesic optimization, we selected the final points of two trajectories for generating a motion from a ring grasp to a spherical grasp. We compared the hyperbolic geodesic with the pullback geodesic. Since the latent space in this experiment captures the taxonomy structure, we incorporated the expected path along the taxonomy nodes into the initialization for the pullback geodesic. Specifically, we concatenated two hyperbolic geodesics: the first connecting the spherical grasp to the origin which represents the root node of the taxonomy graph, and the second connecting the origin to the ring grasp. Starting from this initialization, we applied Riemannian Adam to optimize the geodesic over $200$ steps, using a learning rate of $0.005$ and a spline energy weighting of $\lambda=100$.

\begin{figure*}[t]
    \centering
    \begin{subfigure}[b]{0.495\textwidth}
        \includegraphics[trim={16.0cm 8.5cm 0.0cm 0.0cm},clip,width=0.51\linewidth]{figures/hand_grasps.pdf}
        \verticaldashedsolidblacklinesGrasps
        \includegraphics[trim={15.4cm 0.0cm 4.3cm 12.0cm},clip,width=0.47\linewidth]{figures/hand_grasps.pdf}
        \verticalblackdashedline
        \includegraphics[trim={0.0cm 0.2cm 0.cm 0.2cm},clip,width=0.95\linewidth]{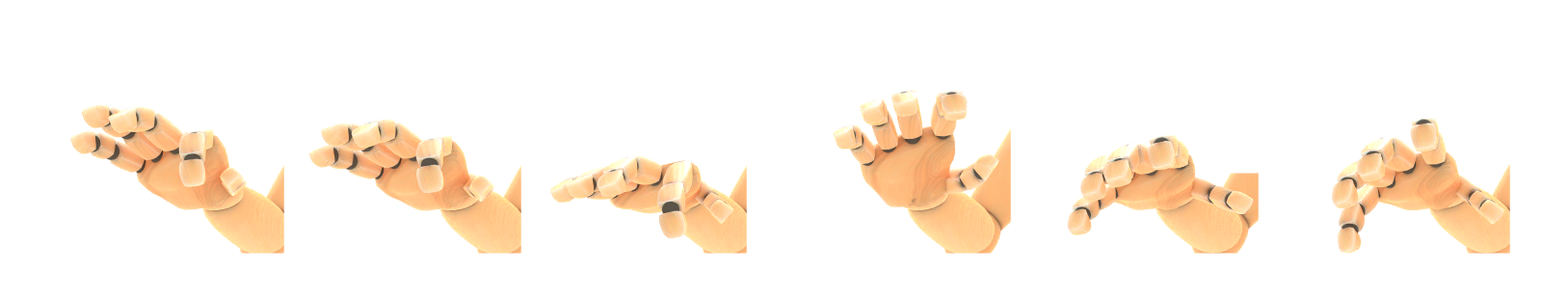}
        \verticalblackline
        \includegraphics[trim={0.0cm 0.2cm 0.cm 0.2cm},clip,width=0.95\linewidth]{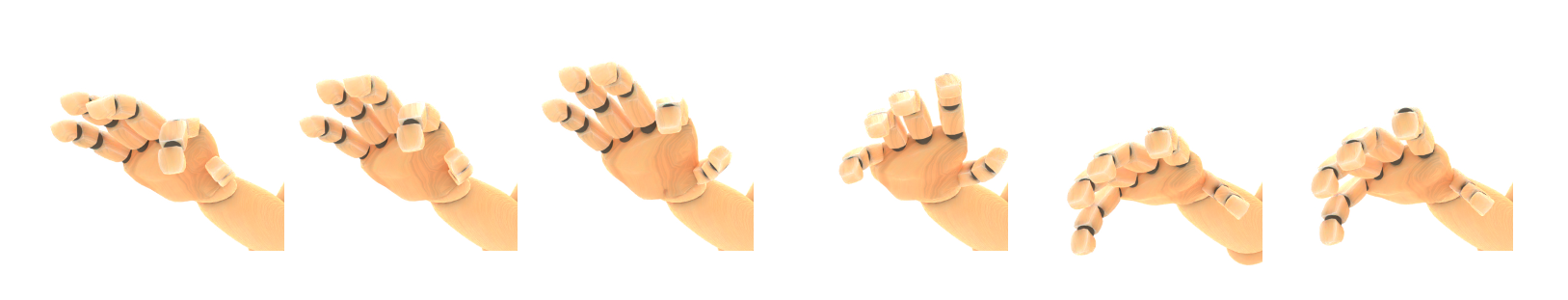}
        \caption{GPHLVM, $\hyperbolic{3}_\mathcal{L}$}
    \end{subfigure}
    \begin{subfigure}[b]{0.495\textwidth}
        \includegraphics[trim={16.0cm 9.4cm 0.0cm 0.0cm},clip,width=0.51\linewidth]{figures/hand_grasps_euclidean.pdf}
        \verticaldashedsolidblacklinesGrasps
        \includegraphics[trim={15.4cm 0.0cm 4.3cm 12.0cm},clip,width=0.47\linewidth]{figures/hand_grasps_euclidean.pdf}
        \verticalblackdashedline
        \includegraphics[trim={0.0cm 0.2cm 0.cm 0.2cm},clip,width=0.95\linewidth]{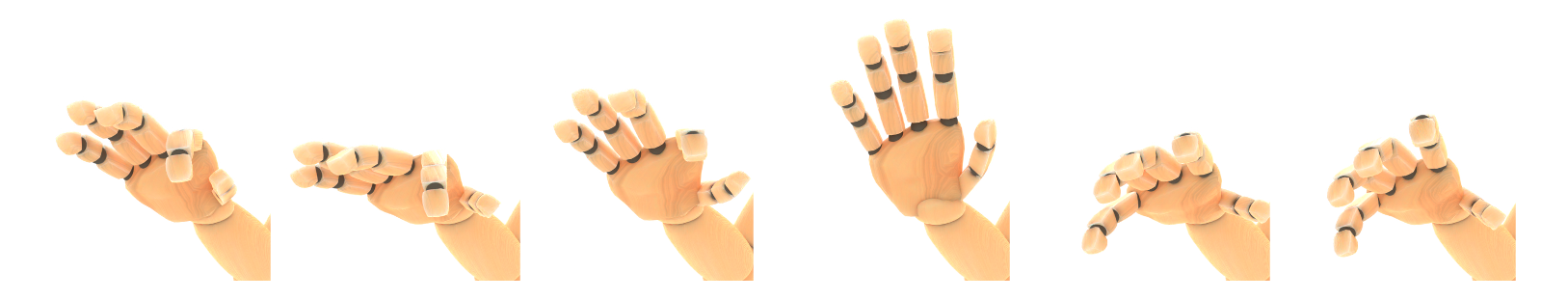}
        \verticalblackline
        \includegraphics[trim={0.0cm 0.2cm 0.cm 0.2cm},clip,width=0.95\linewidth]{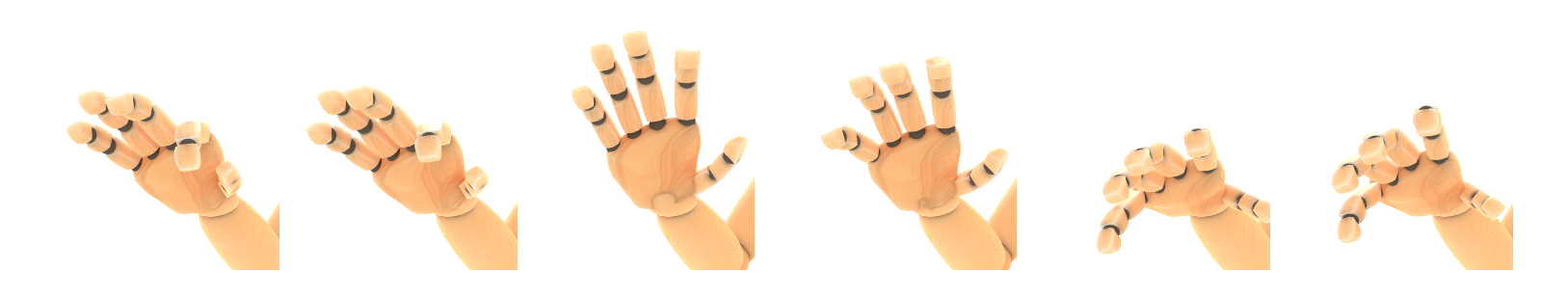}
        \caption{GPLVM, $\mathbb{R}^{3}$}
    \end{subfigure}
    \caption{\emph{Top left}: Embeddings of hand grasps colored according to their corresponding grasp class. The background color represents the pullback metric volume. The base manifold (\blackdashedline) and pullback (\blackline) geodesics correspond to a transition from a ring (\crimsoncircle) to a spherical grasp (\olivecircle).
    \emph{Top right}: Time-series plots of $2$ dimensions of the joint space showing the mean of the decoded geodesics with their uncertainty as a gray envelope. A training trajectory to the spherical grasp (\oliveline) and a reversed training trajectory from the ring grasp (\crimsonline) are included for reference. \emph{Bottom}: Generated hand trajectories from the decoded geodesics.}
    \label{fig:hand_grasps_3D}
\end{figure*}

\begin{figure}[t]
    \centering
    \begin{subfigure}[b]{0.24\textwidth}
        \includegraphics[width=\linewidth]{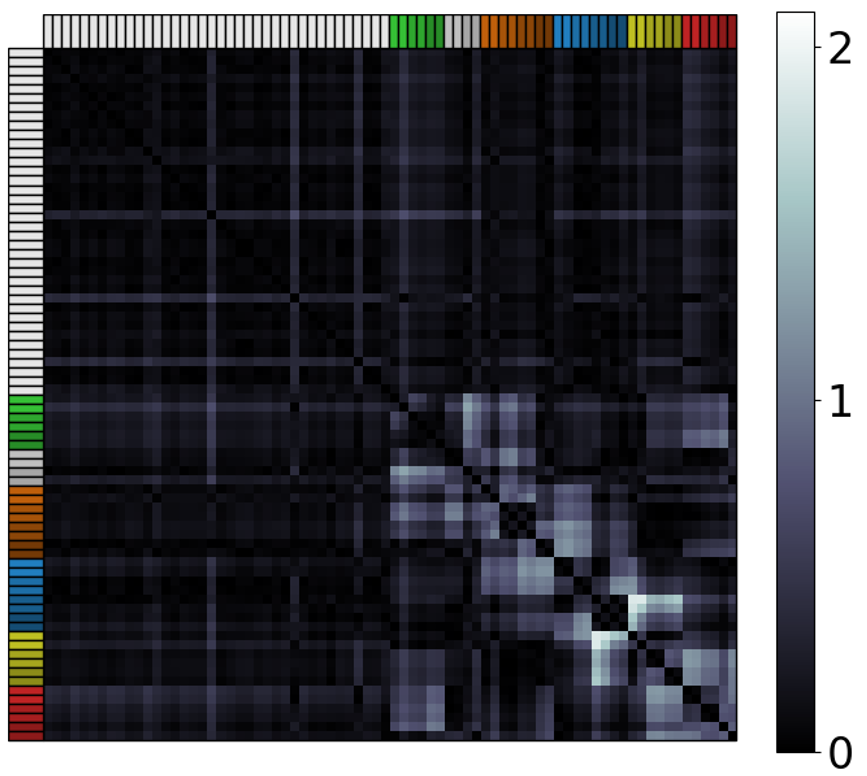}
        \caption{GPHLVM, $\hyperbolic{2}_\mathcal{L}$}
        \label{fig:hand_grasps_error_matrices_H2}
    \end{subfigure}
    \begin{subfigure}[b]{0.24\textwidth}
        \includegraphics[width=\linewidth]{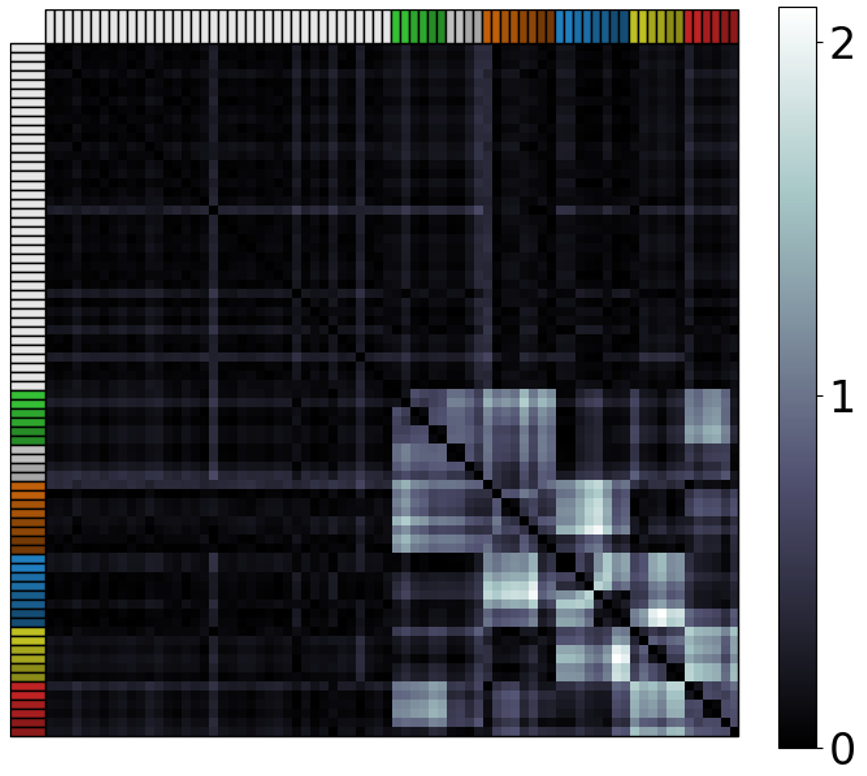}
        \caption{GPLVM, $\mathbb{R}^{2}$}
        \label{fig:hand_grasps_error_matrices_R2}
    \end{subfigure}
    \begin{subfigure}[b]{0.24\textwidth}
        \includegraphics[width=\linewidth]{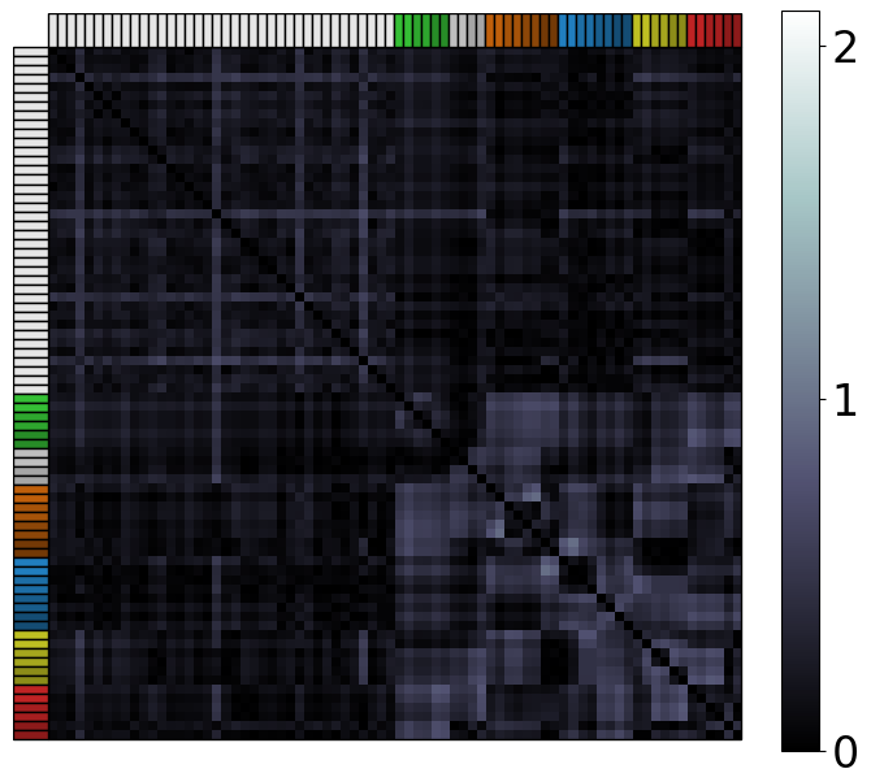}
        \caption{GPHLVM, $\hyperbolic{3}_\mathcal{L}$}
        \label{fig:hand_grasps_error_matrices_H3}
    \end{subfigure}
    \begin{subfigure}[b]{0.24\textwidth}
        \includegraphics[width=\linewidth]{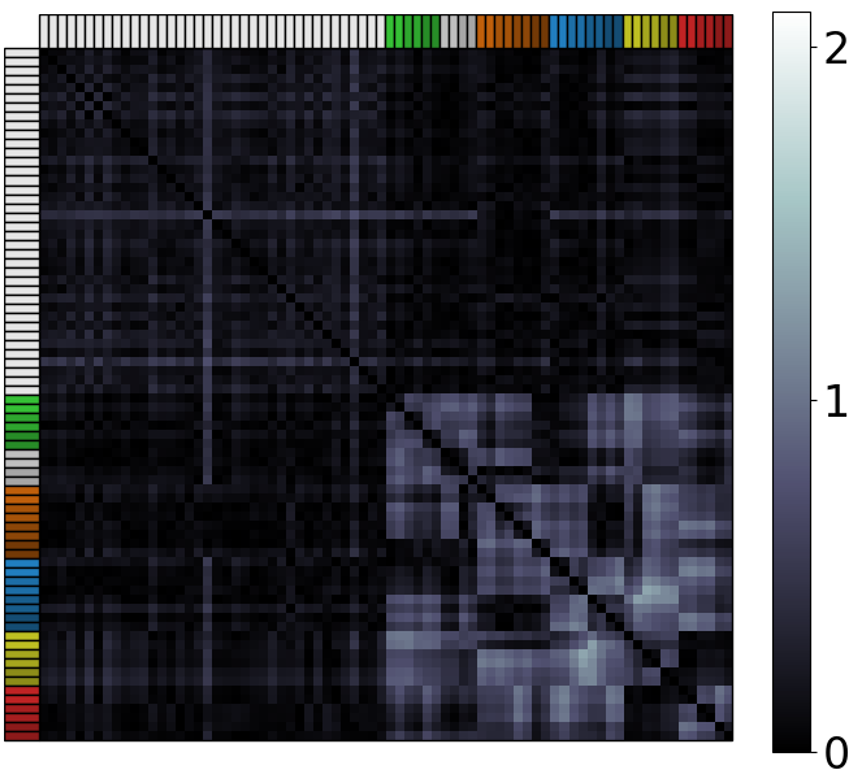}
        \caption{GPLVM, $\mathbb{R}^{3}$}
        \label{fig:hand_grasps_error_matrices_R3}
    \end{subfigure}
    \caption{Pairwise error matrices between geodesic and taxonomy graph distances.}
    \label{fig:hand_grasps_error_matrices}
\end{figure}

\subsection{Additional Hand Grasps Results}
\label{app-sec:hand-grasp-additional-results}
This section provides further insights into the hand grasps experiment described in Sec.~\ref{sec:hand-grasps-generation}. 
First, Fig.~\ref{fig:hand_grasps_3D} shows the 3D hyperbolic and Euclidean latent spaces obtained from the \ac{gphlvm} and \ac{gplvm} of Section~\ref{sec:experiments}, alongside the base manifold and pullback geodesic interpolating from a ring to a spherical grasps and their corresponding decoded hand motions. 

Second, we discuss the stress values provided in Table~\ref{tab:experiments}, which measure the average difference between the hyperbolic distances between embeddings and the corresponding taxonomy graph distances. 
Figure~\ref{fig:hand_grasps_error_matrices} complements the statistics presented in Table~\ref{tab:experiments} and depicts the pairwise error matrices between the graph and manifold distances for the 2D and 3D \acp{gphlvm} and \ac{gplvm}. The left and top bar colors represent the embedding’s taxonomy classes following the color code of Figs.~\ref{fig:hand_grasps} and~\ref{fig:hand_grasps_3D}. The rest pose class (root node of the tree) is represented in white. We observe that, for all models, the pairwise errors between two leaf nodes (i.e., grasps), represented in the bottom-right corner of the plots, are higher than the pairwise errors between rest poses and grasps (i.e., root and leaf nodes), represented in the top-left, top-right, and bottom-left corners. This explains the large variance in the stress values of Table~\ref{tab:experiments}, and is due to the fact that a low error between the root and the leaves simply involves the leaf embeddings to be at the right distance from the origin, where the root is roughly located. Low errors between leaf nodes additionally involve the adequate relative placement of the corresponding embeddings.
It is important to emphasize that the hyperbolic models lead to lower pairwise errors compared to their Euclidean counterparts for all taxonomy classes, as shown by the darker colors of Figs.~\ref{fig:hand_grasps_error_matrices_H2} and~\ref{fig:hand_grasps_error_matrices_H3} compared to Figs.~\ref{fig:hand_grasps_error_matrices_R2} and~\ref{fig:hand_grasps_error_matrices_R3}. Moreover, the 3D hyperbolic model leads to the lowest pairwise errors overall. 

\begin{figure*}[t]
    \centering
    \includegraphics[width=\textwidth]{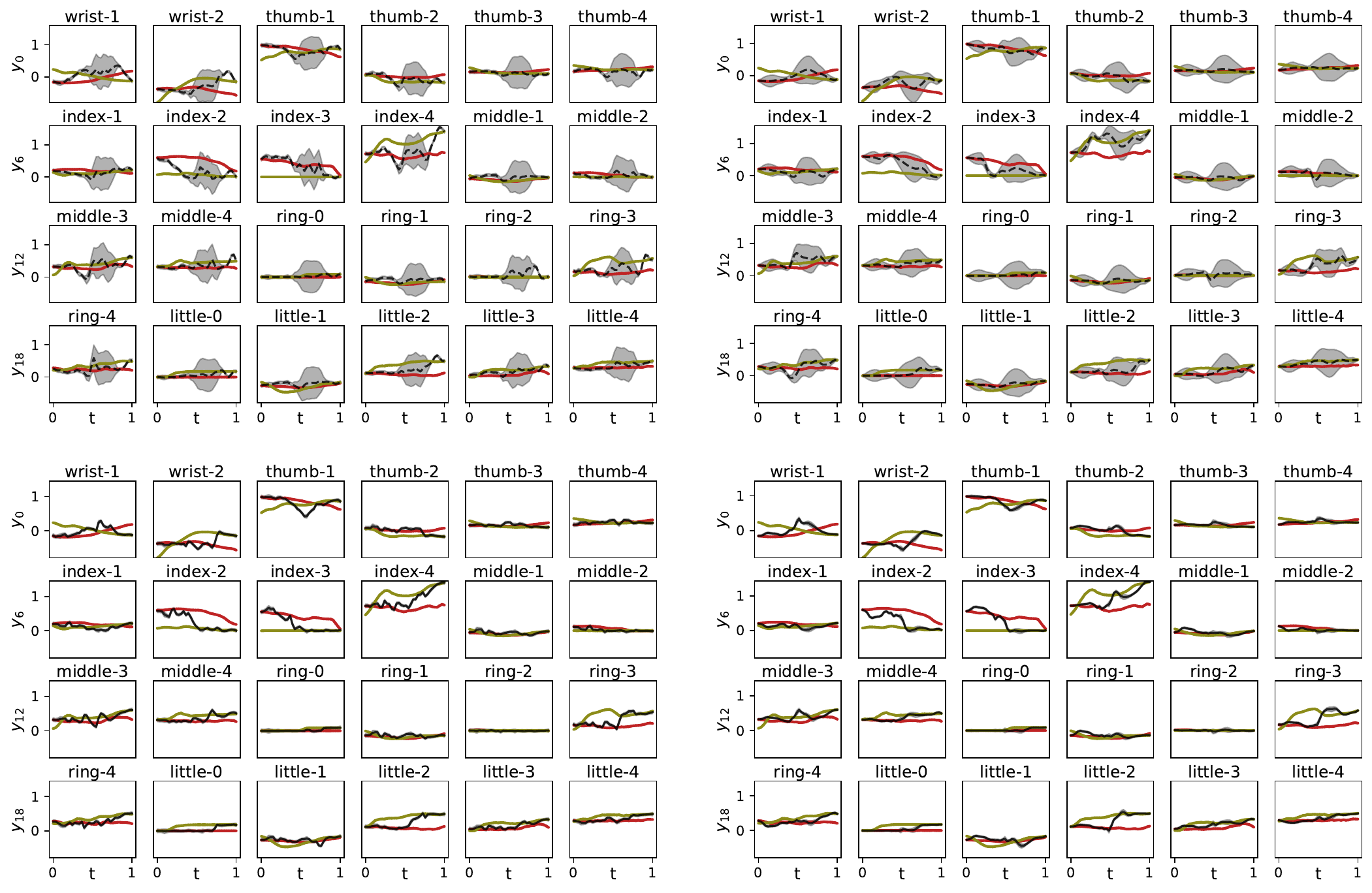}
    \caption{Time-series plots of all $24$ dimensions of the joint space introduced in Fig. \ref{fig:hand_grasps}.
    The left and right columns show the decoded geodesics of the 2D and 3D latent spaces. The top and bottom rows show the mean of the decoded hyperbolic (\blackdashedline) and pullback geodesics (\blackline) and their uncertainty as a gray envelope. For reference, the same training trajectory to the spherical grasp (\oliveline) and reversed training trajectory from the ring grasp (\crimsonline) are included.
    }
    \vspace{-0.2cm}
    \label{fig:hand_grasps_full_predictions}
\end{figure*}

Third, we present additional visualizations of the decoded geodesics. While Figs.~\ref{fig:hand_grasps} and~\ref{fig:hand_grasps_3D} only show $2$ dimensions of the joint space, Fig.~\ref{fig:hand_grasps_full_predictions} depicts the decoded geodesics across all $24$ dimensions, i.e., over the $24$ degrees of freedom (DoF) of the hand model introduced in~\cite{Mandery16:KITmotionDatabase}. This model counts two joints for the wrist, four joints for each finger, and two additional joints for the ring and little fingers to allow a better fist closure.

\begin{figure*}[t]
    \centering
    \includegraphics[width=0.78\textwidth]{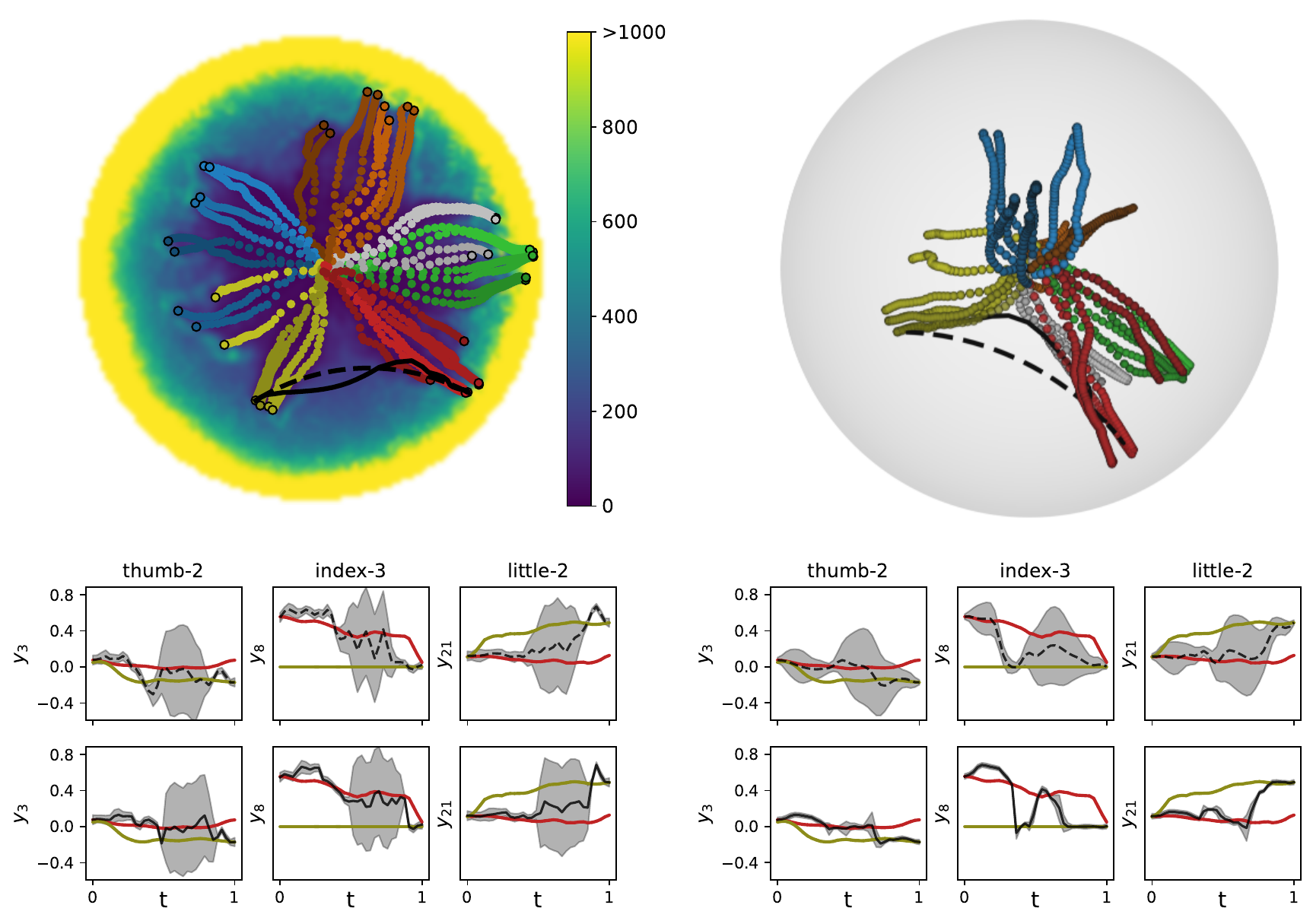}
    \caption{
    Hand grasps transition when initializing the geodesic optimization with the hyperbolic geodesic. As in Fig. \ref{fig:hand_grasps}, a transition from a ring (\crimsoncircle) to a spherical grasp (\olivecircle) is obtained using hyperbolic (\blackdashedline) and pullback (\blackline) geodesics in  $\hyperbolic{2}_\mathcal{L}$ \emph{(left)} and $\hyperbolic{3}_\mathcal{L}$ \emph{(right)}. \emph{Top}: 2D and 3D latent spaces of the trained models.
    \emph{Bottom}: Time-series plots of $3$ dimensions of the joint space for the decoded hyperbolic geodesic (\blackdashedline) and pullback geodesic (\blackline).
    }
    \vspace{-0.2cm}
    \label{fig:hand_grasps_direct_initialization}
\end{figure*}

Finally, we discuss the influence of initialization on optimizing the pullback geodesics. As described in Sec.~\ref{app-sec:hand-grasp-experiment}, we initialize the pullback geodesic with two concatenated hyperbolic geodesics: One from the ring grasp to the origin and another from the origin to the spherical grasp. This initialization incorporates knowledge about the expected path in the taxonomy graph. In contrast, Fig.~\ref{fig:hand_grasps_direct_initialization} shows the optimized pullback geodesics when the optimization is initialized solely with a hyperbolic geodesic from the ring grasp to the spherical grasp.
In the two-dimensional case, we observe that the optimized pullback geodesic path differs significantly from the one shown in Fig.~\ref{fig:hand_grasps}. This discrepancy is likely due to the geodesic optimization getting stuck in a local minimum, which may be a consequence of the Monte Carlo approximation of the 2D hyperbolic SE kernel. Interestingly, the three-dimensional pullback geodesic does not exhibit this issue, further supporting the hypothesis that the Monte Carlo approximation may be responsible since the 3D case does not rely on this approximation. In the 3D case, the optimized pullback geodesic follows a path very similar to the one shown in Fig.~\ref{fig:hand_grasps}.

\subsection{Runtimes}
\label{app-subsec:runtimes}
\begin{table}[t]
    \caption{Average computation times in seconds over $10$ computations of the pullback metric at a random point $\bm{x}$ and of a pullback geodesic between two random points $\bm{x},\bm{y}$. 
    }
    \label{tab:runtimes}
    \begin{center}
    \resizebox{0.55\textwidth}{!}{
    \begin{tabular}{ccc}
    \toprule
        \textbf{Model} & \textbf{Pullback metric} & \textbf{Geodesic} \\
        \midrule
        GPLVM $\euclideanspace^2$         & $0.65 \times 10^{-3} \pm 0.06 \times 10^{-3}$ & $4.14 \pm 0.27$             \\
        GPHLVM $\mathbb{H}^2_\mathcal{L}$ & $44.2 \times 10^{-3} \pm 1.3 \times 10^{-3}$ & $38.08 \pm 0.52$             \\
        \midrule
        GPLVM $\euclideanspace^3$      & $0.63 \times 10^{-3} \pm 0.01 \times 10^{-3}$ & $4.19 \pm 0.23$ \\
        GPHLVM $\mathbb{H}^3_\mathcal{L}$         & $1.77 \times 10^{-3} \pm 0.07 \times 10^{-3}$ & $13.67 \pm 0.33$             \\
    \bottomrule
    \end{tabular}
    }
    \end{center}
\end{table}

Table~\ref{tab:runtimes} shows the runtime measurements for the computation of pullback metrics and pullback geodesics of \ac{gphlvm} and \ac{gplvm} of the hand grasps generation experiment. The implementations are fully developed on Python, and the runtime measurements were taken using a Macbook Pro with Apple M3 Max chip with $14$-core CPU and $36$GB RAM. The main computational burden arises in the \ac{gphlvm} with 2D latent space, which is in contrast with the experiments using a 3D latent space. We attribute this increase of computational cost mainly to the 2D hyperbolic kernel, as also observed in~\cite{Jaquier2024:GPHLVM}. This may be alleviated by reducing the number of samples or via more efficient sampling strategies, as briefly discussed in Section~\ref{sec:conclusions}.

\end{document}